\documentclass[10pt,journal,compsoc]{IEEEtran}
\ifCLASSOPTIONcompsoc
  \usepackage[nocompress]{cite}
\else
  \usepackage{cite}
\fi
\ifCLASSINFOpdf
\else
\fi

\usepackage{hyperref}       
\usepackage{url}            
\usepackage{booktabs}       
\usepackage{amsfonts}       
\usepackage{nicefrac}       
\usepackage{microtype}      

\usepackage{graphicx}
\usepackage{amsmath}
\usepackage{amssymb}
\usepackage{multirow}
\usepackage{bbding}
\usepackage{amstext}

\usepackage[capitalize]{cleveref}
\Crefname{table}{Table}{Tables}
\crefname{section}{Sec.}{Secs.}
\Crefname{section}{Section}{Sections}
\usepackage{paralist}
\usepackage{tabularx}
\usepackage{siunitx}
\newcommand{\myparagraph}[1]{\noindent\textbf{#1}\hspace{0.5em}}

\usepackage{xcolor}

\newcommand{\ie}{\emph{i.e.}}
\newcommand{\eg}{\emph{e.g.}}
\newcommand{\cf}{\emph{cf.}}
\newcommand{\wrt}{\emph{w.r.t.}}
\newcommand{\etal}{\emph{et al.}}

\begin{document}
%
\title{DWDN: Deep Wiener Deconvolution Network\\ for Non-Blind Image Deblurring}
%
%
%
%

\author{Jiangxin Dong, 
        Stefan Roth,
        and~Bernt Schiele,~\IEEEmembership{Fellow,~IEEE}
\IEEEcompsocitemizethanks{\IEEEcompsocthanksitem J.~Dong is with School of Computer Science and Engineering, Nanjing University of Science and Technology, China and also with Department of Computer Vision and Machine Learning, Max Planck Institute for Informatics, Germany. E-mail: dongjxjx@gmail.com
\IEEEcompsocthanksitem S.~Roth is with Department of Computer Science, TU Darmstadt, Germany.\protect\\
E-mail: stefan.roth@visinf.tu-darmstadt.de
\IEEEcompsocthanksitem B.~Schiele is with Department of Computer Vision and Machine Learning, Max Planck Institute for Informatics, Germany.\protect\\
E-mail: schiele@mpi-inf.mpg.de}
}

%
%

\markboth{IEEE Transactions on Pattern Analysis and Machine Intelligence}%
{Dong \MakeLowercase{\textit{et al.}}: Deep Wiener Deconvolution: Wiener Meets Deep Learning for Image Deblurring}
%



\IEEEtitleabstractindextext{%
\begin{abstract}
We present a simple and effective approach for non-blind image deblurring, combining classical techniques and deep learning.
In contrast to existing methods that deblur the image directly in the standard image space, we propose to perform an explicit deconvolution process in a feature space by integrating a classical Wiener deconvolution framework with learned deep features.
A multi-scale cascaded feature refinement module then predicts the deblurred image from the deconvolved deep features, progressively recovering detail and small-scale structures.
The proposed model is trained in an end-to-end manner and evaluated on scenarios with simulated Gaussian noise, saturated pixels, or JPEG compression artifacts as well as real-world images.
Moreover, we present detailed analyses of the benefit of the feature-based Wiener deconvolution and of the multi-scale cascaded feature refinement as well as the robustness of the proposed approach.
Our extensive experimental results show that the proposed \emph{deep Wiener deconvolution network} facilitates deblurred results with visibly fewer artifacts and quantitatively outperforms state-of-the-art non-blind image deblurring methods by a wide margin.
\end{abstract}

}

\maketitle

\IEEEdisplaynontitleabstractindextext

%
\IEEEpeerreviewmaketitle

\IEEEraisesectionheading{\section{Introduction}\label{sec:introduction}}

%
%
%
%
\IEEEPARstart{I}{mage} deblurring is a classical image restoration problem, which has attracted widespread attention~\cite{ChoACM2009,GaoCVPR2019,ZoranICCV2012,AngerICIP2018}, \cite{RenTPAMI2021}.
It is usually formulated as
\begin{equation}
y = x \ast k + n,
\label{eq:blur_model}
\end{equation}
where $y, x, k$, and $n$ denote the blurry input image, the desired clear image, the blur kernel, and image noise, respectively. $\ast$ is the convolution operator. 
Traditional methods usually separate this problem into two phases, blur kernel estimation and image restoration (\ie, non-blind image deblurring).
The goal of non-blind image deblurring is to restore the clear image $x$ from its corrupted observation $y$ given the blur kernel $k$. 
Early non-blind deblurring methods include the Wiener filter~\cite{WienerMIT1949} and the Richardson-Lucy algorithm~\cite{RichardsonOSA1972}.
Later work commonly relied on a probabilistic formulation, most often taking a maximum a-posteriori approach, and much research has been devoted to developing effective image priors \cite{KrishnanNIPS2009,SchmidtCVPR2014,XuECCV2010,ZoranICCV2012} or sophisticated data terms \cite{DongECCV2018,RenPAMI2019}.
However, the optimization problems resulting from such advanced models are difficult to solve, which limits their practical appeal.

\begin{figure*}[!t]
\scriptsize\sffamily
\centering
\begin{tabular}{@{}c@{\hspace{1mm}}c@{\hspace{1mm}}c@{\hspace{1mm}}c@{\hspace{1mm}}c@{\hspace{1mm}}c@{}}
\includegraphics[width = 0.16\linewidth]{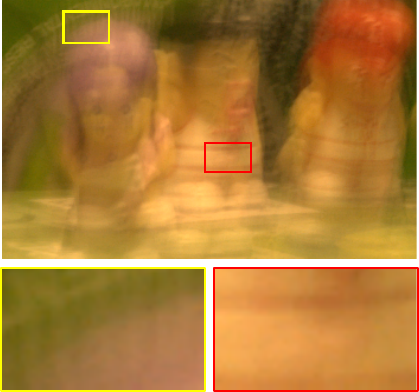}& 
\includegraphics[width = 0.16\linewidth]{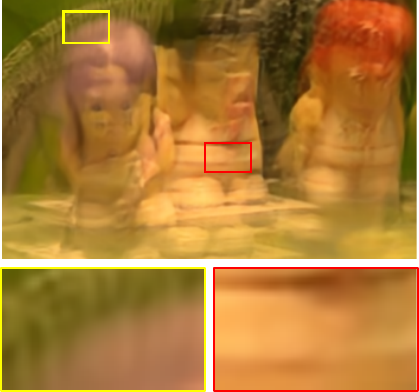}& 
\includegraphics[width = 0.16\linewidth]{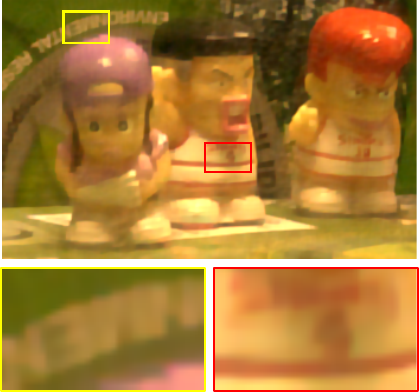}& 
\includegraphics[width = 0.16\linewidth]{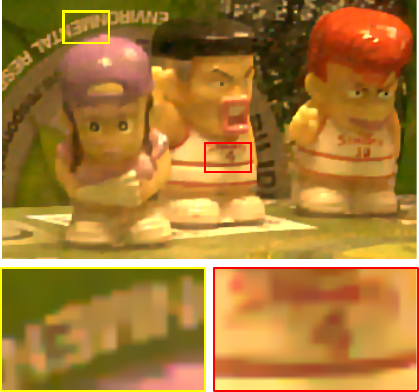}&
\includegraphics[width = 0.16\linewidth]{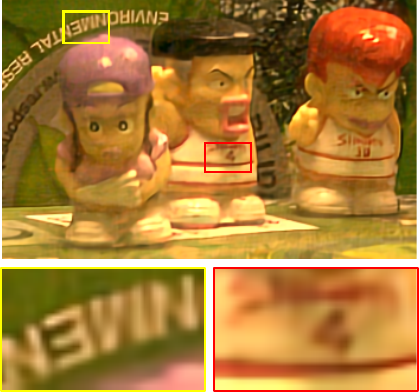}&
\includegraphics[width = 0.16\linewidth]{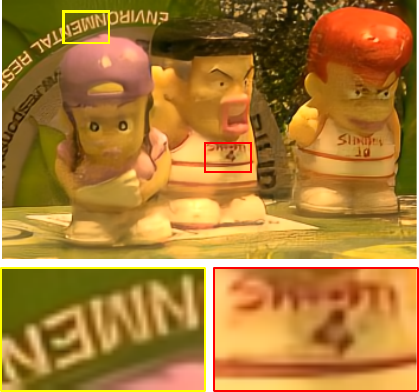} \\
(a) Blurry input & (b) Blind~\cite{ZhangCVPR2019}  & (c) Non-blind~\cite{DongECCV2018} & (d) Non-blind~\cite{PanCVPR2016} & (e) Non-blind~\cite{ZhangCVPR2017} & (f) Our DWDN+ (non-blind)\\
\end{tabular}
\caption{Deblurring results on a real blurry image from~\cite{XuECCV2010}. 
A recent blind image deblurring method~\cite{ZhangCVPR2019}, based on end-to-end trainable networks, does not effectively estimate a clear image. 
With an estimated blur kernel from~\cite{PanCVPR2016}, non-blind image deblurring methods~\cite{DongECCV2018,ZhangCVPR2017,PanCVPR2016} generate better results \emph{(c,d,e)} than the blind method \emph{(b)}. Yet, our DWDN+ method recovers still visibly clearer results \emph{(f)}.
}
\label{fig:motivation1}
\end{figure*}
In recent years, convolutional neural networks (CNNs) have been exploited for image deblurring and shown promising results.
To leverage a-priori knowledge for non-blind image deblurring, one line of work~\cite{SchulerCVPR2013,SonICCP2017,ZhangJiaweiCVPR2017,ZhangCVPR2017} decomposes this problem into image denoising (which is achieved by learned deep models) and image deconvolution\footnote{Note that deconvolution is a particular way of solving image deblurring given the blur model in~\cref{eq:blur_model}.} (performed in the standard image space).
However, the underlying deep models for image denoising are not specifically optimized for the deblurring task, thus not effective for removing deconvolution artifacts and restoring structural detail.
In addition, we observe that performing the deconvolution in the standard image space introduces undesirable artifacts and leads to a loss of fine-scale image detail (\cref{fig:motivation1}(d)).
To remove artifacts, several algorithms use the information extracted by fixed feature extraction operators~\cite{PanICCV2017,ShanACM2008} or discriminatively learned ones~\cite{DongECCV2018,RenPAMI2019} in the deconvolution process.
However, these methods are not effective in finer-scale detail restoration (\cref{fig:motivation1}(c)).
Another line of work adopts generic neural network architectures (\eg, U-Net~\cite{MaoNIPS2016,TaoCVPR2018,ZhangCVPR2019}, GANs~\cite{KupynCVPR2018}) to directly estimate the clear image from the blurry input and achieves reasonable image quality (without the blur kernel being known).
Nevertheless, most of these networks do not perform well compared to established non-blind image deblurring methods if the blur kernel is known, \cf~\cref{fig:motivation1}(b) \emph{vs.}~\cref{fig:motivation1}(c)--(d).
Thus, it is of significant interest to investigate the properties of the deconvolution process and develop effective deep models for non-blind image deblurring.

In this paper, we develop a neural network that integrates a classical deconvolution technique into deep networks for non-blind image deblurring.
In this way, our approach deviates from existing methods~\cite{SonICCP2017,ZhangCVPR2017}, whose deconvolution process is separate from the deep model.
Specifically, we first explore the utility of the classical Wiener deconvolution and propose a \emph{feature-based Wiener deconvolution}. 
Then we embed the feature-based Wiener deconvolution into a deep neural network, which consists of a feature extraction network to provide useful features for the feature-based Wiener deconvolution and a feature refinement network to further refine the deconvolved features for better image reconstruction.
Taken together, we find that the feature-based Wiener deconvolution can better constrain the whole network to be able to learn effective non-blind deblurring.

The contributions of this paper can be summarized as follows:
\emph{(i)} We develop a novel feature-based Wiener deconvolution module that enforces the estimated latent image to coincide with the degradation process in a (deep) feature space.
A detailed analysis demonstrates that this feature-space deconvolution is more effective in suppressing artifacts and recovering fine detail compared to previous methods that conduct the deconvolution in the image space.
Learned deep features further improve the results.
\emph{(ii)} We then propose a multi-scale feature refinement module to restore fine-scale structures from the deconvolved features, facilitating the reconstruction of high-quality images. The whole neural network is trained in an end-to-end manner.
\emph{(iii)} Benefitting from the feature-based Wiener deconvolution, our approach adaptively estimates the noise level from the blurry features, which ensures that training a single instance of the proposed \emph{deep Wiener deconvolution network} is able to handle various levels of noise.
Extensive experiments demonstrate that our approach outperforms existing state-of-the-art methods that require the noise level to be known by a large margin.

This paper is an extension of our earlier conference version \cite{DongNeurIPS2020} with the following key differences.
\emph{(1)} We include a more detailed discussion of the most closely related prior work. 
\emph{(2)} To better refine the deconvolved features and reconstruct final clear images, we improve the multi-scale feature refinement network by implementing cascades of encoder-decoders at each image scale. Specifically, we adopt more encoders and decoders in earlier cascade stages to capture broad contextual information and later fewer ones to focus on details. Residual learning is used in earlier cascade stages to better preserve structural information.
\emph{(3)} To show the effectiveness of our approach over existing state-of-the-art methods, we perform experiments for a wider range of scenarios, including blurry images with various levels of Gaussian noise, saturated pixels, or different degrees of JPEG compression.
\emph{(4)} We conduct more detailed analyses and ablation studies to demonstrate the advantages of the feature-based Wiener deconvolution and the multi-scale cascaded feature refinement and show the robustness of our approach.

\section{Related Work}
\label{sec:related_work}

\myparagraph{Non-learned methods.}
Since non-blind image deblurring is ill-posed, various priors have been proposed to constrain the solution space.
Assuming that natural images have a sparse derivative distribution, total variation~\cite{WangSJIS2008} and hyper-Laplacian priors~\cite{KrishnanNIPS2009,LevinACM2007} have been developed.
To exploit the nonlocal image self-similarity, Danielyan \etal\ \cite{DanielyanTIP2015} develop an iterative deblurring algorithm based on nonlocal patch-wise image modeling~\cite{DabovTIP2007}.
Dong \etal\ \cite{DongTIP2013} present a nonlocally centralized sparse representation by suppressing the sparse coding noise.
Michaeli and Irani~\cite{MichaeliECCV2014} show that the cross-scale patch recurrence property can serve as a strong prior for image deblurring.
Quan \etal\ \cite{QuanJSC2015} construct the self-recurrence of local image structures into wavelet frames.
In addition, Shan \etal\ \cite{ShanACM2008} model the noise distribution by constraining several orders of its derivatives.
Yuan \etal\ \cite{YuanACM2008} propose an inter-scale and intra-scale deconvolution method to recover fine detail while suppressing artifacts.
Common to these hand-crafted priors is that they do not fully exploit the characteristics of clean image data and usually lead to complicated inference problems.

\myparagraph{Classical learning methods.}
To better capture the inherent properties of clear photographic images, various learning methods have been proposed.
Zoran and Weiss \cite{ZoranICCV2012} present a Gaussian mixture prior learned from natural images, which is extended by Sun \etal\ \cite{SunECCV2014} from a single-scale patch prior to a multi-scale formulation.
Roth and Black \cite{RothCSCCVPR2005} learn expressive high-order MRF priors, so called fields of experts.
Schmidt \etal\ \cite{SchmidtPAMI2016} derive a discriminative model based on regression tree fields~\cite{JancsarySBH2012}.
Shrinkage fields~\cite{SchmidtCVPR2014} combine random fields with a discriminatively learned optimization algorithm.
Similarly, Chen \etal\ \cite{ChenCVPR2015} train nonlinear reaction diffusion models by parameterized linear filters and influence functions.
To better model the image reconstruction error, \cite{DongECCV2018,RenPAMI2019} learn a set of filters and penalty functions to model the data term.
While achieving decent image quality, these learned priors or penalty functions usually require the design of sophisticated numerical algorithms.

%
\begin{figure*}[!t]\scriptsize
\centering
\includegraphics[width = 1.0\linewidth]{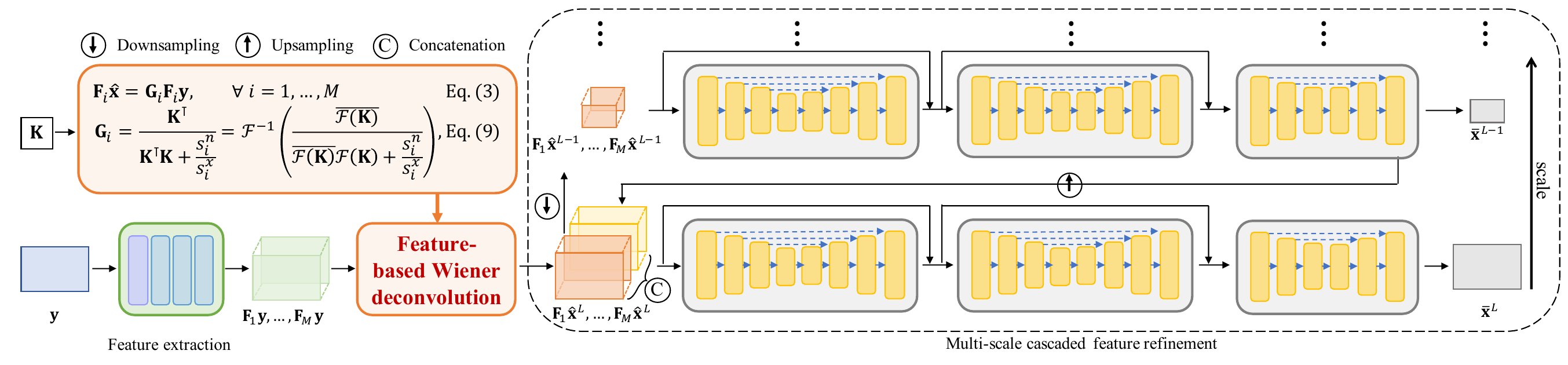}
\vspace{-5mm}
\caption{Deep Wiener deconvolution network. While previous work mostly relies on a deconvolution in the image space, our neural network first extracts useful feature information from the blurry input image and then conducts an explicit Wiener deconvolution in the (deep) feature space through~\cref{eq:deconvolution,eq:wiener_filter}. A multi-scale cascaded encoder-decoder network progressively restores clear images, with fewer artifacts and finer detail. The whole network is trained in an end-to-end manner.
}
\label{fig: pipeline}
\end{figure*}

\myparagraph{Deep learning methods.}
More recently, deep learning has been used for non-blind image deblurring.
Xu \etal\ \cite{XuNIPS14} use the singular value decomposition of the pseudo-inverse kernel to initialize the network parameters. 
However, their method needs fine-tuning for every kernel.
To overcome this, Ren \etal\ \cite{RenNIPS2018} propose a generalized deep CNN by exploiting the low-rank property of pseudo-inverse kernels.
In addition, several methods~\cite{SchulerCVPR2013,ZhangJiaweiCVPR2017,ZhangCVPR2017,MeinhardtICCV2017,RomanoSIAM2017,RyuICML2019} decompose non-blind deblurring into two individual subproblems, \ie, image denoising and image deconvolution.
Zhang \etal\ \cite{ZhangJiaweiCVPR2017} first deconvolve the blurry image in the image space and then use a fully convolutional network to learn image gradients to further guide the final deconvolution.
\cite{SchulerCVPR2013,ZhangCVPR2017} train a set of CNNs to improve the denoising accuracy.
Meinhardt \etal\ \cite{MeinhardtICCV2017} learn a fixed denoising network as a proximal operator in the primal-dual hybrid gradient method \cite{ChambolleJMIV2011}.
These methods achieve reasonable image quality, but separately designing these two subproblems -- denoising and deconvolution -- will make the approach not be fully optimized for image deblurring.
Dong \etal\ \cite{DongPAMI2019} propose a denoising-based network, which contains several denoiser modules interleaved with back-projection modules that are jointly optimized.
Kruse \etal\ \cite{KruseICCV2017} generalize discriminative FFT-based deconvolution by using CNN-based regularization and propose a boundary adjustment method.
Kobler \etal\ \cite{KoblerPR2017} introduce variational networks learned by minimizing a parametrized energy using proximal incremental methods \cite{BertsekasMP2011}.
Eboli \etal\ \cite{EboliECCV2020} adopt the Richardson fixed-point iteration method \cite{KelleySIAM1995}, where the parameters are learned by embedding the preconditioner into a CNN.
Gong \etal\ \cite{GongTNNLS2020} incorporate deep neural networks into a gradient descent scheme. 
Quan \etal\ \cite{QuanTNNLS2021} introduce a CNN-based image prior defined in the Gabor domain combined with an unrolled optimization scheme.
However, most of these approaches implement the deconvolution process in the standard image space. Hence, artifacts remain and fine-scale detail is lost, as pointed out by~\cite{DongECCV2018,PanICCV2017,ShanACM2008}.

\myparagraph{Other related work.}
Son and Lee \cite{SonICCP2017} use a Wiener deconvolution as a preprocessing step and then develop residual networks with long and short skip-connections to remove artifacts caused by the Wiener deconvolution. Their network is able to remove artifacts but does not effectively preserve image detail. Thus, they adopt an additional postprocessing step to remedy this problem.
Different from~\cite{SonICCP2017}, we show that the Wiener deconvolution implemented in the standard image space is less effective for artifact suppression and detail restoration and thus propose a novel feature-based Wiener deconvolution.
In addition, the three steps in \cite{SonICCP2017} are separately employed, where the latter steps are used to compensate the errors introduced by the former ones while the former ones do not take into account the latter ones.
In contrast, our approach incorporates a feature-based Wiener deconvolution with a feature refinement in an end-to-end architecture, which facilitates each module benefitting from one another and does not require any pre-/postprocessing.

While most prior work assumes the noise level to be known, noise-blind methods offer an alternative. 
Schmidt \etal\ \cite{SchmidtCVPR2011} develop a sampling-based Bayesian method with integrated noise estimation.
Jin \etal\ \cite{JinCVPR2017} introduce a sampling-free approach using a smoothed generalization of the $0$-$1$ loss.
Bigdeli \etal\ \cite{BigdeliNIPS2017} formulate image deblurring as a Bayes estimator and propose a deep mean-shift prior built on denoising autoencoders to learn the density of natural images.
Nan \etal\ \cite{NanCVPR2020} integrate the estimation of the noise level and the quantification of the uncertainty \wrt~the image prior in a variational expectation maximization framework.
Different from~\cite{JinCVPR2017,SchmidtCVPR2011,BigdeliNIPS2017,NanCVPR2020}, we explore an explicit Wiener deconvolution step in a deep feature space and adaptively estimate the noise level from the blurry features, ensuring that training a single model is able to handle various levels of noise.

Most aforementioned methods work well for blurry images with Gaussian noise, yet are not as effective when applied to blurry inputs with significant outliers (\eg, saturated pixels and/or JPEG compression artifacts).
To handle impulsive noise, Bar \etal\ \cite{BarIJCV2006} derive an $\ell_1$ norm-based data term based on the Laplacian distribution.
Cho \etal\ \cite{ChoICCV2011} analyze how various types of outliers violate the commonly used blur assumption to explicitly classify outliers and exclude them from the deblurring process.
To deal with saturated pixels, Whyte \etal\ \cite{WhyteIJCV2014} modify the Richardson-Lucy algorithm to properly treat the saturated pixels and prevent ringing artifacts.
Xu \etal\ \cite{XuNIPS2014} perform deconvolution based on the kernel separability theorem using deep neural networks, where the nonlinear terms and high-dimensional structure make the network more expressive and robust to outliers.
Benefitting from specific domain knowledge, these methods work well on certain types of outliers but cannot be easily extended to other types.
In contrast, our approach does not rely on particular assumptions of noise or outliers and can adaptively learn to handle various types of noise or outliers from the training data (\eg, Gaussian noise, saturated pixels, and/or JPEG compression artifacts).

\begin{figure*}[!t]
\scriptsize\sffamily
\centering
\begin{tabular}{@{}c@{\hspace{1mm}}c@{\hspace{1mm}}c@{\hspace{1mm}}c@{\hspace{1mm}}c@{}}
\includegraphics[width = 0.24\linewidth]{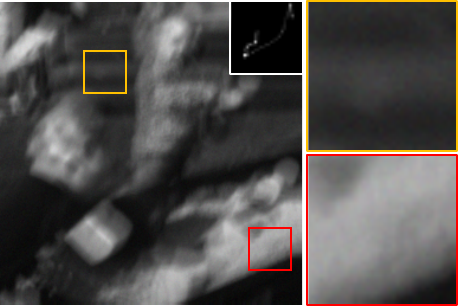}& 
\includegraphics[width = 0.24\linewidth]{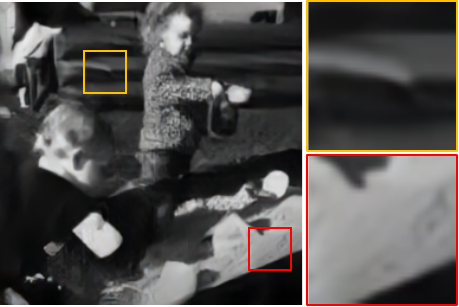}& 
\includegraphics[width = 0.24\linewidth]{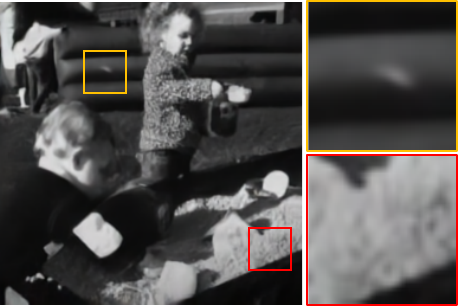}& 
\includegraphics[width = 0.24\linewidth]{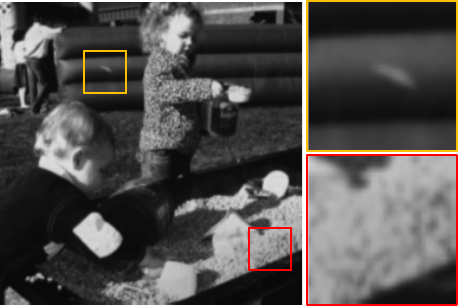} \\
(a) Blurry input & (b) Standard image space  & (c) Deep feature space & (d) Ground truth\\
\end{tabular}
\caption{
\emph{(a)} Blurry image and blur kernel. \emph{(b)} and \emph{(c)} show the results of methods that perform the deconvolution in the standard image space and a deep feature space, respectively. \emph{(d)} Ground truth. The deblurred result \emph{(c)} from the proposed approach contains fewer artifacts and much more detail in the yellow and red boxes than those in \emph{(b)}.
}
\label{fig:motivation}
\end{figure*}

\section{Deep Wiener Deconvolution Network}
\label{sec:proposed_method}
Our goal is to develop an effective non-blind image deblurring approach to restore high-quality images with few artifacts and fine detail.
Specifically, the proposed deep Wiener deconvolution network contains a feature-based Wiener deconvolution module and a multi-scale cascaded feature refinement module, which are trained in an end-to-end manner.
\Cref{fig: pipeline} summarizes the model architecture.

\subsection{Feature-based Wiener deconvolution}
\label{ssec:feature_deconvolution}
To design a neural network specifically for the task of image deblurring, we propose to embed an explicit Wiener deconvolution step into our network.
However, only using the intensity information in the standard image space for the deconvolution is not effective for artifact removal and detail restoration as pointed out by~\cite{DongECCV2018,PanICCV2017,ShanACM2008}. 
Thus, we propose a \emph{feature-based Wiener deconvolution module} to better constrain the deconvolution process with useful feature information from the blurry input.

Let $\{f_i\}_{i=1}^M$ denote a set of linear filters, which are used to extract useful feature information from the blurry input.
By convolving both sides of~\cref{eq:blur_model} with $\{f_i\}$, we can obtain the relationship between $y, x$, and $k$ in the feature space, owing to the properties of convolution, as
\begin{equation}
\textbf{F}_i \textbf{y} = \textbf{K} \textbf{F}_i \textbf{x} + \textbf{F}_i \textbf{n}, \quad\forall i=1,\dots,M,
\label{eq:blur_model_feature}
\end{equation}
where $\textbf{F}_i, \textbf{K}, \textbf{y}, \textbf{x}$, and $\textbf{n}$ denote the matrix/vector forms of $f_i, k, y, x$, and $n$.

The goal of our feature-based Wiener deconvolution module is to explicitly deconvolve the blurry features $\{\textbf{F}_i \textbf{y}\}$ from~\cref{eq:blur_model_feature} by finding a set of \emph{feature-based Wiener deconvolution operators} $\{\textbf{G}_i\}$ so that we can obtain the latent features as
\begin{equation}
\textbf{F}_i \hat{\textbf{x}} = \textbf{G}_i \textbf{F}_i \textbf{y}, \quad\forall i=1,\dots,M,
\label{eq:deconvolution}
\end{equation}
where $\hat{\textbf{x}}$ is the latent clear image sought after.
In order to obtain latent features that are close to the ground-truth clear features, for each $i$, we need to minimize the mean squared error~\cite{WienerMIT1949}
\begin{subequations}
\begin{align}
e_i  &= \mathbb{E} \big( | \textbf{F}_i \textbf{x} - \textbf{F}_i \hat{\textbf{x}} |^2 \big) = \mathbb{E} \big( | \textbf{F}_i \textbf{x} - \textbf{G}_i \textbf{F}_i \textbf{y} |^2 \big)\\
     &= \mathbb{E} \big( | \textbf{F}_i \textbf{x} - \textbf{G}_i (\textbf{K} \textbf{F}_i \textbf{x} + \textbf{F}_i \textbf{n}) |^2 \big)\\
     &= (1-\textbf{G}_i \textbf{K}) (1-\textbf{G}_i \textbf{K})^\top \mathbb{E} \big( | \textbf{F}_i \textbf{x} |^2 \big) \notag\\
     &~~~~- (1-\textbf{G}_i \textbf{K}) \textbf{G}_i^\top \mathbb{E} \big( \textbf{F}_i \textbf{x} (\textbf{F}_i\textbf{n})^\top \big)    \label{eq:e1}\\
     &~~~~- (1-\textbf{G}_i \textbf{K})^\top \textbf{G}_i \mathbb{E} \big( (\textbf{F}_i \textbf{x})^\top \textbf{F}_i\textbf{n} \big) + \textbf{G}_i \textbf{G}_i^\top \mathbb{E} \big( |\textbf{F}_i\textbf{n}|^2 \big), \notag
\end{align}
\end{subequations}
where $\mathbb{E}$ denotes the expectation.
Assuming the noise to be independent from the latent clear image and having zero mean, we can derive that
\begin{equation}
\mathbb{E} \big( \textbf{F}_i \textbf{x} (\textbf{F}_i\textbf{n})^\top \big) = \mathbb{E} \big( \textbf{F}_i \textbf{x} \big) \mathbb{E} \big( (\textbf{F}_i\textbf{n})^\top \big) = 0
\end{equation}
and
\begin{equation}
\mathbb{E} \big( (\textbf{F}_i \textbf{x})^\top \textbf{F}_i\textbf{n} \big) = \mathbb{E} \big( (\textbf{F}_i \textbf{x})^\top \big) \mathbb{E} \big( \textbf{F}_i\textbf{n} \big) = 0.
\end{equation}
We denote $\mathbb{E} \big( | \textbf{F}_i \textbf{x} |^2 \big)$ and $\mathbb{E} \big( |\textbf{F}_i\textbf{n}|^2 \big)$ by $s_i^x$ and $s_i^n$ and then rewrite~\cref{eq:e1} as
\begin{equation}
e_i = \big( 1-\textbf{G}_i \textbf{K} \big) \big( 1-\textbf{G}_i \textbf{K} \big)^\top s_i^x + \textbf{G}_i \textbf{G}_i^\top s_i^n.
\label{eq:e2}
\end{equation}

To minimize $e_i$, we compute the derivative of~\cref{eq:e2} with respect to $\textbf{G}_i$ and set it to zero:
\begin{equation}
 \big( \textbf{K}^\top\textbf{K} s_i^x + s_i^n \big) \textbf{G}_i - \textbf{K}^\top s_i^x = 0.
\end{equation}
Then we can obtain the feature-based Wiener deconvolution operator $\textbf{G}_i$ as
\begin{equation}
\textbf{G}_i = \frac{\textbf{K}^\top}{\textbf{K}^\top\textbf{K} + \frac{s_i^n}{s_i^x}} = \mathcal{F}^{-1} \Bigg(\frac{\overline{\mathcal{F}(\textbf{K})}}{\overline{\mathcal{F}(\textbf{K})}\mathcal{F}(\textbf{K}) + \frac{s_i^n}{s_i^x}}\Bigg),
\label{eq:wiener_filter}
\end{equation}
where $\mathcal{F}$ denotes the discrete Fourier transform and $\overline{\mathcal{F}(\textbf{K})}$ is the complex conjugate of $\mathcal{F}(\textbf{K})$.
Thus, we can obtain the latent feature $\{\textbf{F}_i \hat{\textbf{x}} \}$ by~\cref{eq:deconvolution,eq:wiener_filter}.

To extract useful feature information from the blurry input, we can choose common derivative operators (\eg, first- and higher-order derivatives) or discriminatively learned linear filters.
In general, the input of the feature-based Wiener deconvolution module is the degraded image and the outputs are the deconvolved latent features $\{\textbf{F}_i \hat{\textbf{x}} \}$.
As our analysis in~\cref{ssec:Effect of the feature-based Wiener deconvolution} will show, the Wiener deconvolution step is more effective when combining more and more useful feature information.

In addition, we can obtain more powerful feature extractors $\{\textbf{F}_i\}$ in~\cref{eq:deconvolution} using deep neural networks~\cite{MildenhallCVPR2018,NiklausCVPR2017,DiamondArXiv2017}.
While deep feature extractors are not linear, as assumed by \cref{eq:blur_model_feature}, they are locally linear \cite{MontufarNIPS2014,LeeICLR2019}.
Hence, we apply \cref{eq:blur_model_feature} regardless; remaining errors can be compensated by the feature refinement (\cref{ssec:multi-scale_feature_refinement}).
We directly estimate the blurry features $\{\textbf{F}_i \textbf{y} \}$ given as input to the deconvolution $\textbf{G}_i$ and use a feature extraction network with one convolutional layer followed by three residual blocks~\cite{HeCVPR2016}. 
By embedding the feature-based Wiener deconvolution into an end-to-end network, the feature extraction network can learn more useful features for deconvolution, \cf~\cref{ssec:Effect of the feature-based Wiener deconvolution}.

We estimate $s_i^x$ as the standard deviation of the blurry feature $\textbf{F}_i \textbf{y}$ and estimate $s_i^n$ as the variance of the difference between the blurry feature $\textbf{F}_i \textbf{y}$ and the mean-filtered result of $\textbf{F}_i \textbf{y}$.
$s_i^n$ can thus approximately capture the noise level of each blurry feature. A more detailed analysis is included in \cref{ssec:Robustness of the proposed DWDN approach}.
Benefitting from the feature-based Wiener deconvolution and the multi-scale feature refinement in~\cref{ssec:multi-scale_feature_refinement}, our network is able to handle blurry images with various noise levels.

To intuitively illustrate the effect of the feature-based Wiener deconvolution module, we compare performing the Wiener deconvolution in the standard image space and in a deep feature space in~\cref{fig:motivation}(b) and (c).
For fair comparison, both methods use the same multi-scale feature refinement (without the cascaded architecture) in~\cref{ssec:multi-scale_feature_refinement} to reconstruct the final results.
\Cref{fig:motivation} shows that deconvolving the blurry image in a deep feature space is much more effective at yielding a clear image with fewer artifacts (\eg, the wall in the yellow boxes of \cref{fig:motivation}) and finer textures (\eg, the sand in the red boxes of \cref{fig:motivation}).

\subsection{Multi-scale cascaded feature refinement}
\label{ssec:multi-scale_feature_refinement} 
To restore high-quality images from the deconvolved latent features, we develop a multi-scale feature refinement module in a coarse-to-fine manner.
Specifically, we first build a pyramid $\{ \{ \textbf{F}_i \hat{\textbf{x}}^l \}_{l=1}^L\}$ of the full-resolution latent features $\{ \textbf{F}_i \hat{\textbf{x}} \}$ from \cref{eq:deconvolution} using bicubic downsampling with a scale factor of $2$.
Then we can estimate the clear image $\bar{\textbf{x}}^l$ at each scale by
\begin{equation}
\begin{split}
\bar{\textbf{x}}^l&=\mathcal{N} (\textbf{h}^l), \\ 
\textbf{h}^l&=
\begin{cases}
\mathcal{C}\big(\textbf{F}_1 \hat{\textbf{x}}^l,\dots, \textbf{F}_M \hat{\textbf{x}}^l\big) & \text{if }{l=1,}\\
\begin{aligned}[b]
  \mathcal{C}\big(&\textbf{F}_1 \hat{\textbf{x}}^l,\dots, \textbf{F}_M \hat{\textbf{x}}^l,\\
  &\mathcal{N}_{-1}(\textbf{F}_1 \hat{\textbf{x}}^{l-1},\dots, \textbf{F}_M \hat{\textbf{x}}^{l-1})\uparrow\big)
\end{aligned} & \text{if }{l=2,\ldots,L,}
\end{cases}\raisetag{39pt}
\end{split}
\label{eq:multi-scale}
\end{equation}
where $\mathcal{N}$ denotes a network for the feature refinement module, $\mathcal{N}_{-1}$ denotes the network~$\mathcal{N}$ without the last layer, $\mathcal{C}$ is the concatenation, and $\uparrow$ is the upsampling operation.

\begin{figure}[!t]\scriptsize
\centering
\includegraphics[width = 0.7\linewidth]{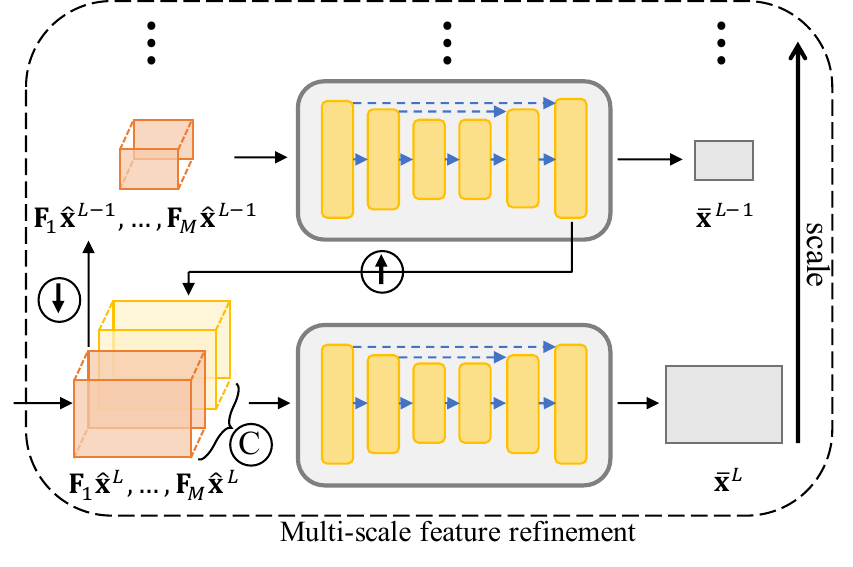}
\vspace{-3mm}
\caption{Multi-scale feature refinement in our conference version \cite{DongNeurIPS2020}.
}
\label{fig: pipeline2}
\vspace{-2mm}
\end{figure}

In our previous conference paper \cite{DongNeurIPS2020}, we use an encoder-decoder architecture for the network~$\mathcal{N}$ as shown in \cref{fig: pipeline2}, whose effect is further discussed in \cref{ssec:Effect of the feature refinement}.
We denote our approach in the conference paper \cite{DongNeurIPS2020} as \emph{DWDN}.
To better refine the latent features and reconstruct the final clear image, we here propose a multi-scale cascaded feature refinement module as shown in \cref{fig: pipeline}.
At each image scale, we employ cascades of encoder-decoder pairs, which adopt more encoders and decoders in earlier cascade stages to capture broad context, as well as fewer ones in later stages to focus more on image detail.
To avoid losing fine-scale image information with more encoders and decoders, we adopt residual learning in the earlier cascade stages.
Our improved approach is denoted as \emph{DWDN+}.
As shown in \cref{eq:multi-scale}, the number of input channels in the feature refinement network for scale $l=1$ and $l>1$ differs.
Hence, our feature refinement network shares its parameters across all scales except for the first encoder block of the first cascade.

%

\begin{table*}[!t]
  \caption{Quantitative comparison to state-of-the-art methods on the dataset of Levin \etal\ \cite{LevinCVPR2009} with Gaussian noise of $1\%, 3\%$, and $5\%$.}
  \vspace{-2mm}
  \label{tab:levin_dataset}
  \centering
  \scriptsize
  \begin{tabularx}{\textwidth}{@{}c@{$\;\;$}X@{}c@{\hspace{1.0mm}}c@{\hspace{1.0mm}}c@{\hspace{1.0mm}}c@{\hspace{1.0mm}}c@{\hspace{1.0mm}}c@{\hspace{1.0mm}}c@{\hspace{1.0mm}}c@{\hspace{1.0mm}}c@{\hspace{1.0mm}}c@{\hspace{1.0mm}}c@{\hspace{1.0mm}}c@{\hspace{1.0mm}}c@{}}
    \toprule
Noise level& & DMPHN~\cite{ZhangCVPR2019} & EPLL~\cite{ZoranICCV2012}  &MLP~\cite{SchulerCVPR2013} &CSF~\cite{SchmidtCVPR2014} &LDT~\cite{DongECCV2018} &FCN~\cite{ZhangJiaweiCVPR2017} &IRCNN~\cite{ZhangCVPR2017} &FDN~\cite{KruseICCV2017}&FNBD~\cite{SonICCP2017} &CPCR~\cite{EboliECCV2020} &RGDN~\cite{GongTNNLS2020} & DWDN~\cite{DongNeurIPS2020} & DWDN+\\
    \midrule
    \multirow{2}{*}{$1\%$}
&PSNR (dB)&25.95  &34.06  &32.08  &31.90  &31.53  &33.22  &34.33  &34.05  &34.81  &33.75  &33.96  &36.90  &\bf37.27 \\
&SSIM     &0.7918 &0.9310 &0.8884 &0.9024 &0.8977 &0.9267 &0.9210 &0.9335 &0.9398 &0.9283 &0.9395 &0.9614 &\bf0.9634\\
    \midrule
    \multirow{2}{*}{$3\%$}
&PSNR (dB)&25.78  &29.09  &27.00  &28.01  &28.39  &29.49  &30.04  &29.77  &30.63  &29.58  &29.71  &32.77  &\bf33.09 \\
&SSIM     &0.7814 &0.8460 &0.7016 &0.8013 &0.8052 &0.8599 &0.8156 &0.8583 &0.8658 &0.8566 &0.8662 &0.9179 &\bf0.9217\\
    \midrule
    \multirow{2}{*}{$5\%$}
&PSNR (dB)&25.19  &26.54  &25.38  &26.32  &26.70  &27.72  &28.51  &27.94  &27.93  &27.77  &27.45  &30.77  &\bf31.08 \\
&SSIM     &0.7579 &0.7785 &0.6330 &0.7427 &0.7468 &0.8142 &0.7762 &0.8139 &0.7759 &0.8108 &0.7889 &0.8857 &\bf0.8906\\
    \bottomrule
  \end{tabularx}
\end{table*}

\begin{table*}[!t]
  \caption{Quantitative comparison to state-of-the-art methods on the dataset of~\cite{MartinICCV2001} with Gaussian noise of $1\%, 3\%$, and $5\%$.}
  \vspace{-2mm}
  \label{tab:bsds_dataset}
  \centering
  \scriptsize
  \begin{tabularx}{\textwidth}{@{}c@{$\;\;$}X@{}c@{\hspace{1.0mm}}c@{\hspace{1.0mm}}c@{\hspace{1.0mm}}c@{\hspace{1.0mm}}c@{\hspace{1.0mm}}c@{\hspace{1.0mm}}c@{\hspace{1.0mm}}c@{\hspace{1.0mm}}c@{\hspace{1.0mm}}c@{\hspace{1.0mm}}c@{\hspace{1.0mm}}c@{\hspace{1.0mm}}c@{}} 
    \toprule
Noise level& & DMPHN~\cite{ZhangCVPR2019} & EPLL~\cite{ZoranICCV2012}  &MLP~\cite{SchulerCVPR2013} &CSF~\cite{SchmidtCVPR2014} &LDT~\cite{DongECCV2018} &FCN~\cite{ZhangJiaweiCVPR2017} &IRCNN~\cite{ZhangCVPR2017} &FDN~\cite{KruseICCV2017}&FNBD~\cite{SonICCP2017} &CPCR~\cite{EboliECCV2020} &RGDN~\cite{GongTNNLS2020} & DWDN~\cite{DongNeurIPS2020} & DWDN+\\
    \midrule
    \multirow{2}{*}{$1\%$}
&PSNR (dB)&24.21  &29.81  &28.47  &29.00  &28.20  &29.51  &30.63  &29.93  &30.92  &29.41  &29.51  &31.74  &\bf32.02 \\
&SSIM     &0.6572 &0.8385 &0.7977 &0.8230 &0.7922 &0.8339 &0.8645 &0.8555 &0.8799 &0.8270 &0.8616 &0.8938 &\bf0.9001\\
    \midrule
    \multirow{2}{*}{$3\%$}
&PSNR (dB)&24.04  &26.28  &25.62  &26.33  &26.24  &26.92  &27.18  &27.23  &27.44  &26.70  &27.06  &28.58  &\bf28.73 \\
&SSIM     &0.6538 &0.6996 &0.6505 &0.7096 &0.7018 &0.7346 &0.7219 &0.7505 &0.7618 &0.7188 &0.7620 &0.8040 &\bf0.8113\\
    \midrule
    \multirow{2}{*}{$5\%$}
&PSNR (dB)&23.72  &24.66  &24.01  &24.93  &24.90  &25.45  &25.65  &25.93  &25.49  &25.44  &25.33  &27.29  &\bf27.41 \\
&SSIM     &0.6255 &0.6276 &0.5619 &0.6428 &0.6358 &0.6771 &0.6640 &0.6943 &0.6589 &0.6640 &0.6688 &0.7573 &\bf0.7651 \\
    \bottomrule
  \end{tabularx}
\vspace{-2mm}
\end{table*}

\begin{table*}[!t]
  \caption{Quantitative comparison to state-of-the-art methods on the dataset of~\cite{SunICCP2013} with Gaussian noise of $1\%, 3\%$, and $5\%$.}
  \vspace{-2mm}
  \label{tab:sun_dataset}
  \centering
  \scriptsize
  \begin{tabularx}{\textwidth}{@{}c@{$\;\;$}X@{}c@{\hspace{1.0mm}}c@{\hspace{1.0mm}}c@{\hspace{1.0mm}}c@{\hspace{1.0mm}}c@{\hspace{1.0mm}}c@{\hspace{1.0mm}}c@{\hspace{1.0mm}}c@{\hspace{1.0mm}}c@{\hspace{1.0mm}}c@{\hspace{1.0mm}}c@{\hspace{1.0mm}}c@{\hspace{1.0mm}}c@{}}
    \toprule
Noise level& & DMPHN~\cite{ZhangCVPR2019} & EPLL~\cite{ZoranICCV2012}  &MLP~\cite{SchulerCVPR2013} &CSF~\cite{SchmidtCVPR2014} &LDT~\cite{DongECCV2018} &FCN~\cite{ZhangJiaweiCVPR2017} &IRCNN~\cite{ZhangCVPR2017} &FDN~\cite{KruseICCV2017}&FNBD~\cite{SonICCP2017} &CPCR~\cite{EboliECCV2020} &RGDN~\cite{GongTNNLS2020} & DWDN~\cite{DongNeurIPS2020} & DWDN+\\
    \midrule
    \multirow{2}{*}{$1\%$}
&PSNR (dB)&24.75  &32.48  &31.47  &31.52  &30.52  &32.36  &33.57  &32.63  &31.22  &32.00  &31.25  &34.05 &\bf34.62 \\ 
&SSIM     &0.7111 &0.8815 &0.8535 &0.8622 &0.8399 &0.8853 &0.8977 &0.8887 &0.8860 &0.8750 &0.8869 &0.9225&\bf0.9303\\ 
    \midrule
    \multirow{2}{*}{$3\%$}
&PSNR (dB)&26.03  &28.40  &26.23  &28.10  &28.14  &29.04  &29.07  &29.09  &29.70  &28.78  &28.53  &30.71 &\bf30.98 \\ 
&SSIM     &0.7068 &0.7589 &0.6035 &0.7408 &0.7374 &0.7841 &0.7430 &0.7803 &0.8040 &0.7764 &0.7992 &0.8447&\bf0.8533\\ 
    \midrule
    \multirow{2}{*}{$5\%$}
&PSNR (dB)&25.80  &26.78  &24.65  &26.62  &26.71  &27.67  &27.64  &27.75  &27.63  &27.51  &26.93  &29.30 &\bf29.49 \\ 
&SSIM     &0.6925 &0.6975 &0.5198 &0.6735 &0.6694 &0.7340 &0.6884 &0.7319 &0.7010 &0.7277 &0.7161 &0.8006&\bf0.8083 \\ 
    \bottomrule
  \end{tabularx}
\vspace{-2mm}
\end{table*}

\myparagraph{Loss function.}
To better regularize the network, we apply an $\ell_1$ loss function at each scale $l$.
The final loss function is given as
\begin{equation}
\mathcal{L}  = \sum_{l=1}^{L} \frac{\gamma_l}{N_l} \big\| \bar{\textbf{x}}^l - \textbf{x}^l \big\|_1,
\label{eq:loss}
\end{equation}
where $\textbf{x}^l$ is the downsampled ground-truth image using bicubic interpolation for the scale $l$, $\{ \gamma_l \}$ are the weights for each scale, and $N_l$ is the number of elements in $\textbf{x}^l$ for normalization.
%

\section{Experimental Results}
\label{sec:experimental_results}
Next, we describe the implementation details of our proposed approach in \cref{ssec:Implementation details}.
Then, we evaluate our method against the state of the art on blurry images with simulated Gaussian noise (\cref{ssec:Results with simulated blur and Gaussian noise}), saturated pixels (\cref{ssec:Results with simulated blur and saturated pixels}), and JPEG compression (\cref{ssec:Results with simulated blur and JPEG compression}), as well as on real-world images (\cref{ssec:Results with real blur}).
More experimental results are included in the supplemental material.

\begin{figure*}[!t]
\scriptsize\sffamily
\centering
\begin{tabular}{@{}c@{\hspace{1mm}}c@{\hspace{1mm}}c@{\hspace{1mm}}c@{\hspace{1mm}}c@{\hspace{1mm}}c@{}}
\includegraphics[width = 0.16\linewidth]{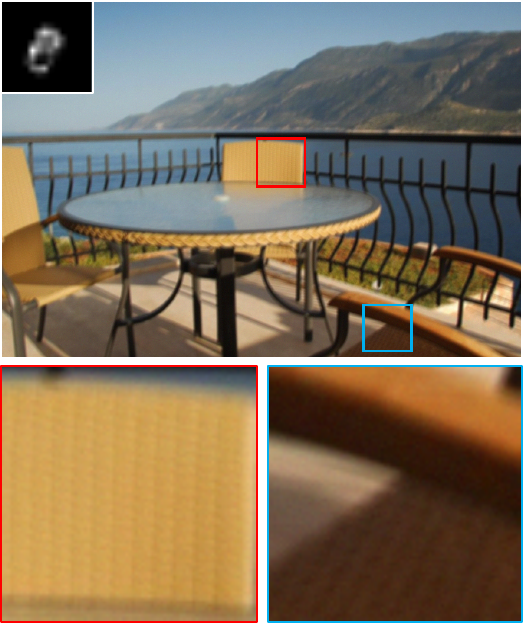}& 
\includegraphics[width = 0.16\linewidth]{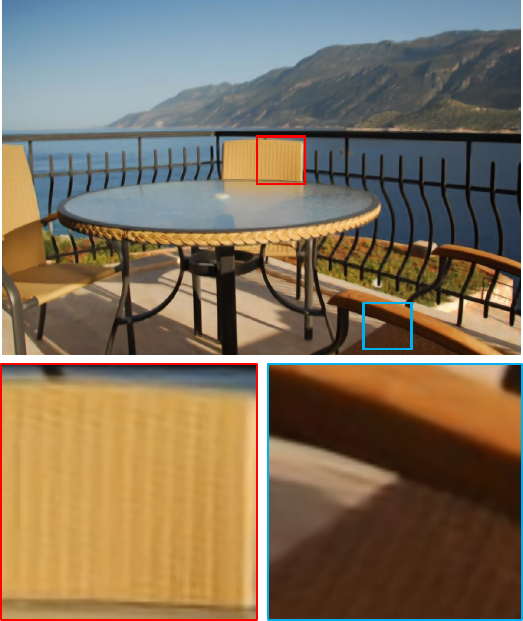}& 
\includegraphics[width = 0.16\linewidth]{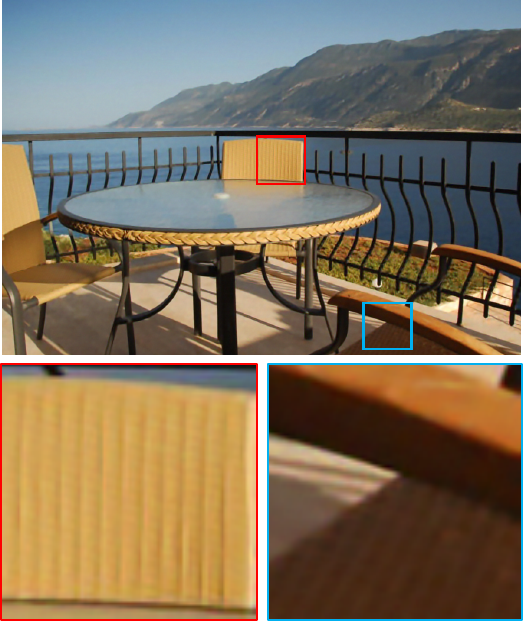}& 
\includegraphics[width = 0.16\linewidth]{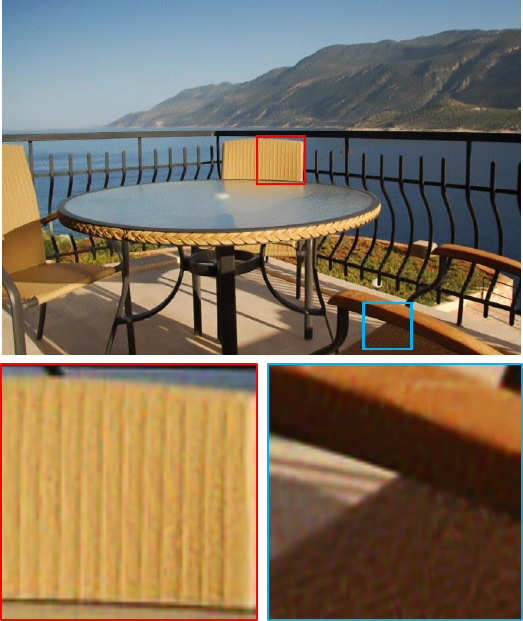}& 
\includegraphics[width = 0.16\linewidth]{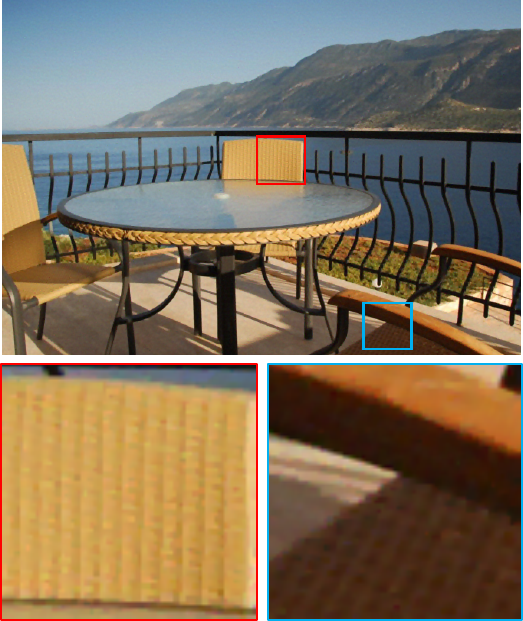} & 
\includegraphics[width = 0.16\linewidth]{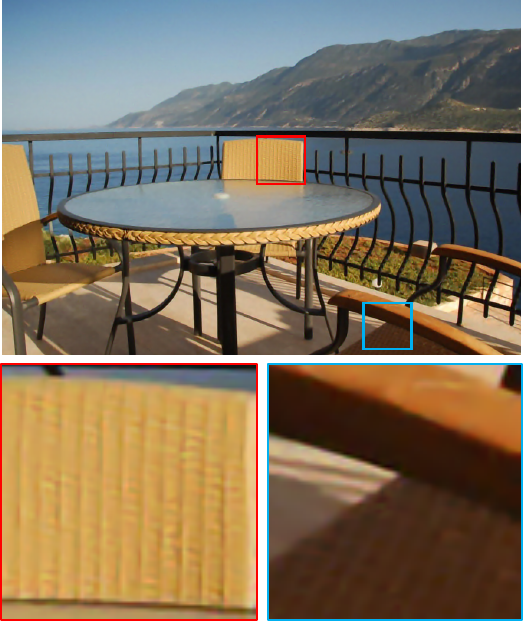}\\
(a) Blurry input & (b) DMPHN~\cite{ZhangCVPR2019} & (c)  EPLL~\cite{ZoranICCV2012} & (d) MLP~\cite{SchulerCVPR2013} & (e) CSF~\cite{SchmidtCVPR2014} & (f) FCN \cite{ZhangJiaweiCVPR2017} \\[1mm]
\includegraphics[width = 0.16\linewidth]{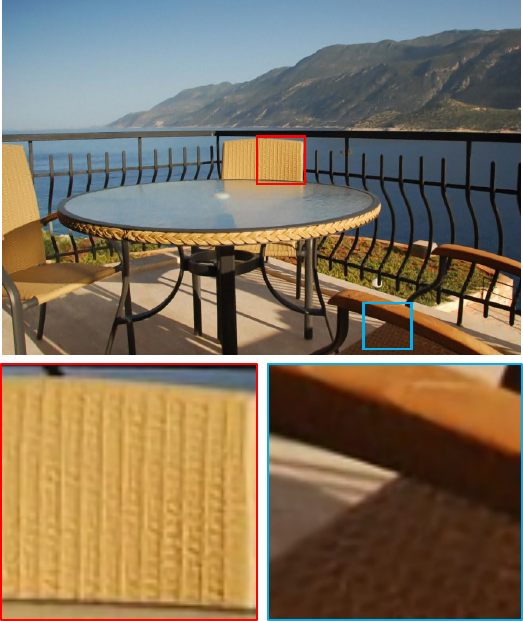}& 
\includegraphics[width = 0.16\linewidth]{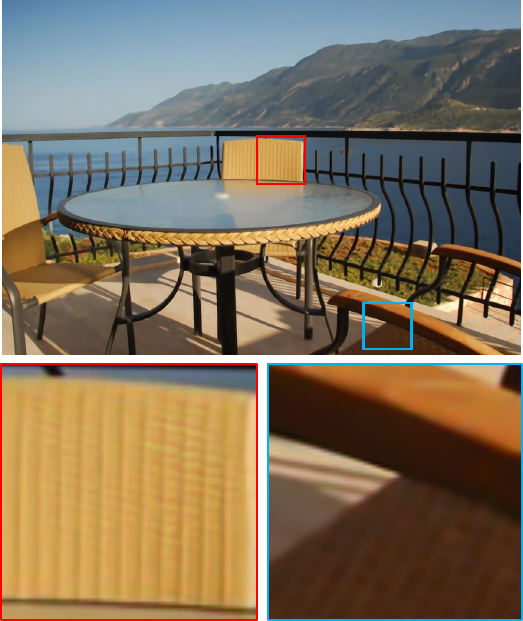}& 
\includegraphics[width = 0.16\linewidth]{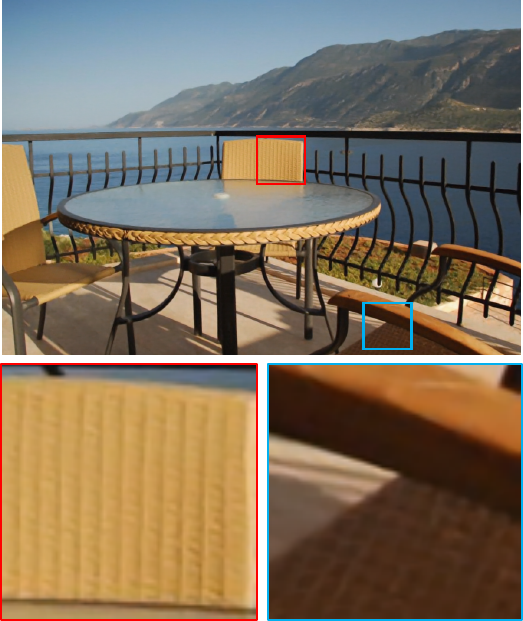}& 
\includegraphics[width = 0.16\linewidth]{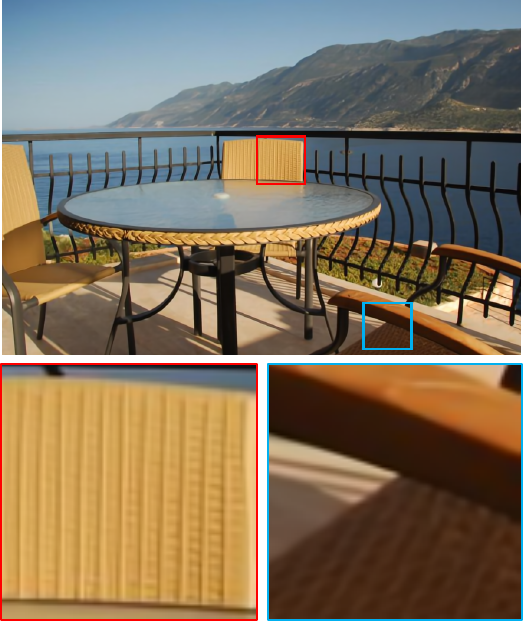}& 
\includegraphics[width = 0.16\linewidth]{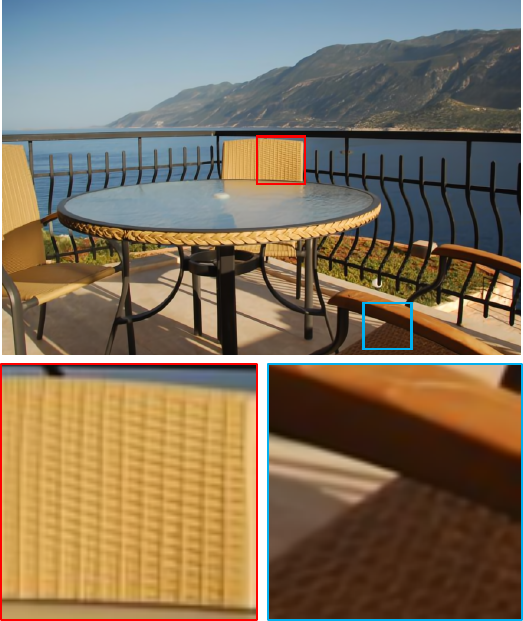} & 
\includegraphics[width = 0.16\linewidth]{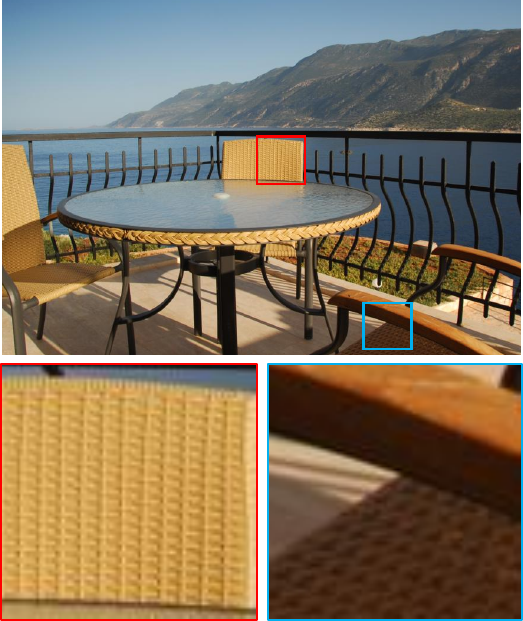}\\
(g) FNBD~\cite{SonICCP2017} & (h) CPCR \cite{EboliECCV2020} & (i)RGDN \cite{GongTNNLS2020} & (j) DWDN  & (k) DWDN+ & (l) Ground truth\\
\end{tabular}
\caption{
Example with simulated blur ($1\%$ noise level) from the dataset of~\cite{SunICCP2013}. The result obtained by~\cite{SchulerCVPR2013} has severe artifacts in \emph{(d)}. For other methods, small-scale structures and detail are over-smoothed as shown in the red and blue boxes of \emph{(c)} and \emph{(e)--(i)}. Compared to existing methods, our DWDN approach can effectively preserve finer detail as shown in \emph{(j)}. Our improved multi-scale cascaded feature refinement (DWDN+) can make small-scale structures even clearer and sharper as shown in \emph{(k)}.
}
\label{fig:visual_results_levin5}
\vspace{-2mm}
\end{figure*}

\subsection{Implementation details}
\label{ssec:Implementation details}
Balancing effectiveness and efficiency, we use a total of two scales and three cascades in the multi-scale cascaded feature refinement module.
We empirically use $M=16$ features and set $\gamma_l=1$. 
In both training and test phases, we pad the blurry features before applying the Wiener deconvolution.
Specifically, we replicate $ks$ pixels of the boundary of each blurry feature on all sides, where $ks$ denotes the size of the blur kernel.
For training the network parameters, we adopt the Adam optimizer~\cite{Kingma2014} with default parameters.
The batch size is set to $8$.
The learning rate is initialized as $10^{-4}$, which is halved every $200$ epochs.
PyTorch code and trained models are available at our project page.\footnote{\href{https://gitlab.mpi-klsb.mpg.de/jdong/dwdn}{\nolinkurl{gitlab.mpi-klsb.mpg.de/jdong/dwdn}}}
As pointed out in \cite{TaiPAMI2013,PanCVPR2016}, the non-linear camera response function can be addressed before the deconvolution steps, hence we do not consider these factors in our experiments.
\subsection{Results with simulated blur and Gaussian noise}
\label{ssec:Results with simulated blur and Gaussian noise}
\myparagraph{Training dataset.} We collect a training dataset including \num{400} images from the Berkeley segmentation~\cite{MartinICCV2001} and \num{4744} images from the Waterloo Exploration~\cite{MaTIP2016} datasets.
Specifically, we generate the training dataset as is common in the non-blind deblurring literature~\cite{ZhangJiaweiCVPR2017,ZhangCVPR2017}.
We randomly crop patches of $256 \times 256$ pixels from each clear image.
Then we synthesize realistic kernels~\cite{SchmidtPAMI2016} of random sizes in the range from $13 \times 13$ to $35 \times 35$ pixels and convolve each clear image patch with a blur kernel.
Finally, we add Gaussian noise to the blurry images with noise levels ranging from $0$ to $5\%$. 
We train a single model for \emph{all noise levels} and \emph{without fine-tuning} for each test scenario.

\myparagraph{Test datasets.} We first use the popular benchmark test datasets of Levin \etal\ \cite{LevinCVPR2009} and Sun \etal\ \cite{SunICCP2013} to evaluate our approach.
These two datasets contain 4 and 80 clear images, respectively, while adopting the same 8 blur kernels from~\cite{LevinCVPR2009}.
Next, the proposed method is evaluated on a test dataset generated using $100$ clear images from the dataset of~\cite{MartinICCV2001} and $100$ synthetic blur kernels from~\cite{SchmidtPAMI2016}.
We evaluate all methods with different Gaussian noise levels ($1\%, 3\%$, and $5\%$ on the datasets of \cite{LevinCVPR2009,SunICCP2013,MartinICCV2001}).
The training and all test datasets do not overlap.

\begin{table*}[!t]
  \caption{Quantitative comparison to state-of-the-art methods on a dataset with saturated pixels (see text for details).}
  \vspace{-2mm}
  \label{tab:sat_dataset}
  \centering
  \scriptsize
  \begin{tabularx}{\textwidth}{@{}Xc@{\hspace{1.8mm}}c@{\hspace{1.8mm}}c@{\hspace{1.8mm}}c@{\hspace{1.8mm}}c@{\hspace{1.8mm}}c@{\hspace{1.8mm}}c@{\hspace{1.8mm}}c@{\hspace{1.8mm}}c@{\hspace{1.8mm}}c@{\hspace{1.8mm}}c@{\hspace{1.8mm}}c@{\hspace{1.8mm}}c@{}}
    \toprule
& EPLL~\cite{ZoranICCV2012}  &MLP~\cite{SchulerCVPR2013} &CSF~\cite{SchmidtCVPR2014} &LDT~\cite{DongECCV2018} &FCN~\cite{ZhangJiaweiCVPR2017} &IRCNN~\cite{ZhangCVPR2017} &FDN~\cite{KruseICCV2017}&FNBD~\cite{SonICCP2017} &RGDN~\cite{GongTNNLS2020} & Whyte~\cite{WhyteIJCV2014} & Cho~\cite{ChoICCV2011} & DWDN~\cite{DongNeurIPS2020} & DWDN+\\
    \midrule
PSNR (dB)&29.78  &28.60  &29.28 &30.52 &29.14  &29.92 &28.20 &27.48 &28.61 &28.14 &33.02 &33.70 &\bf34.14\\
SSIM     &0.8950 &0.8652 &0.8931 &0.9167 &0.8789 &0.9089&0.8560&0.8739&0.8919&0.8824&0.9388&0.9535&\bf0.9564\\
    \bottomrule
  \end{tabularx}
  \vspace{-1mm}
\end{table*}

\begin{figure*}[!t]
\scriptsize\sffamily
\centering
\begin{tabular}{@{}c@{\hspace{1mm}}c@{\hspace{1mm}}c@{\hspace{1mm}}c@{\hspace{1mm}}c@{\hspace{1mm}}c@{}}
\includegraphics[width = 0.16\linewidth]{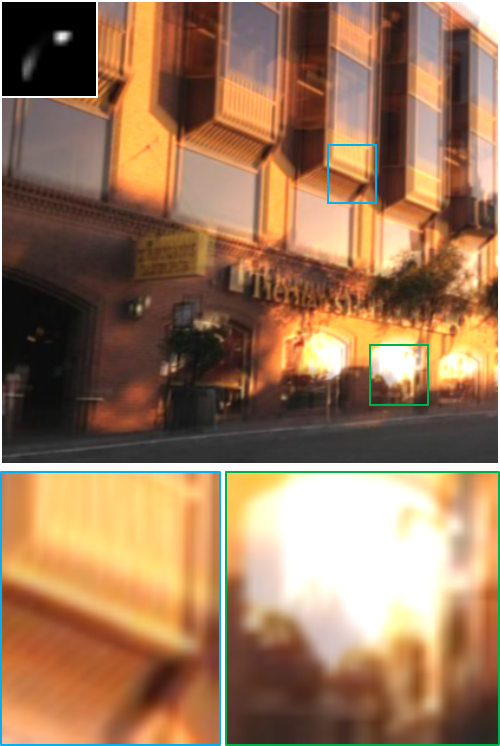}& 
\includegraphics[width = 0.16\linewidth]{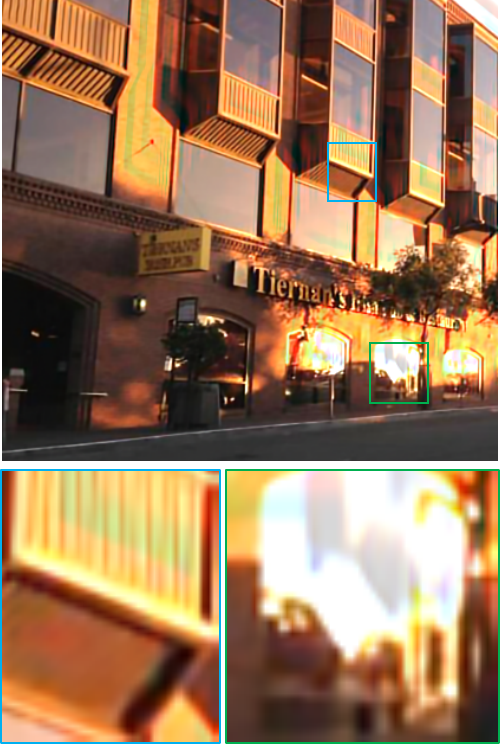}& 
\includegraphics[width = 0.16\linewidth]{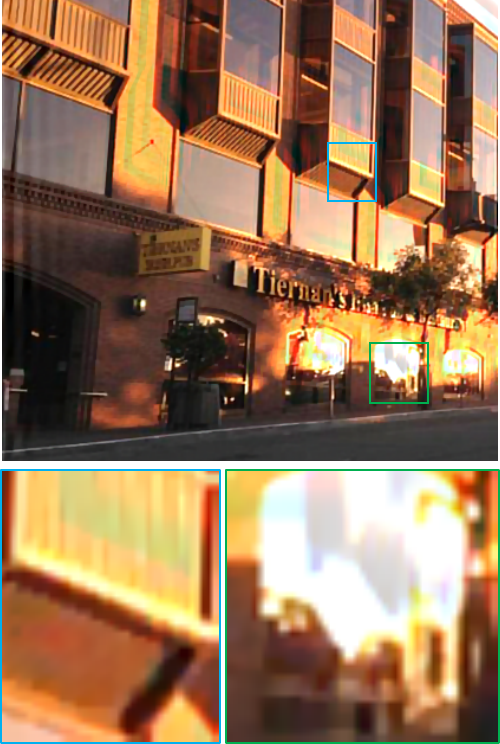}& 
\includegraphics[width = 0.16\linewidth]{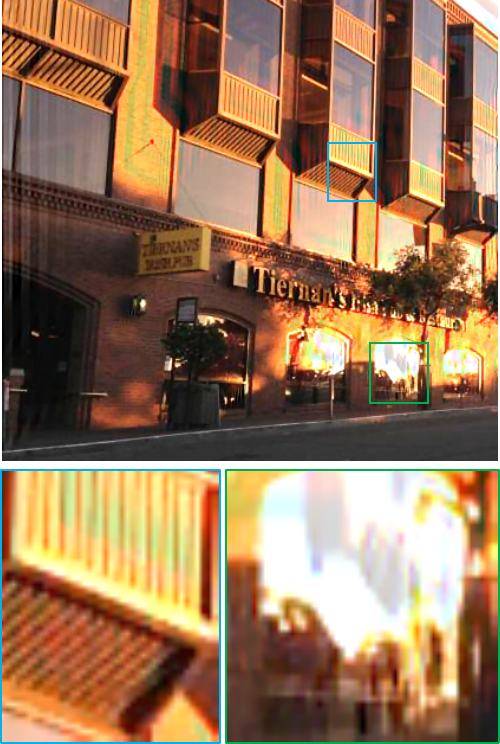}& 
\includegraphics[width = 0.16\linewidth]{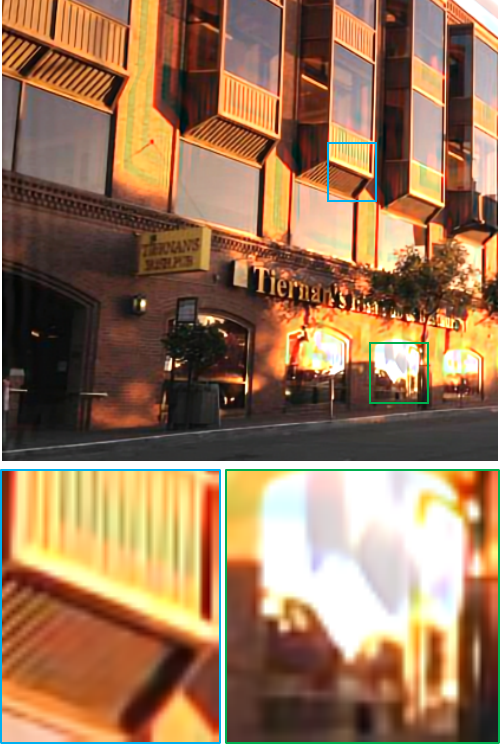} & 
\includegraphics[width = 0.16\linewidth]{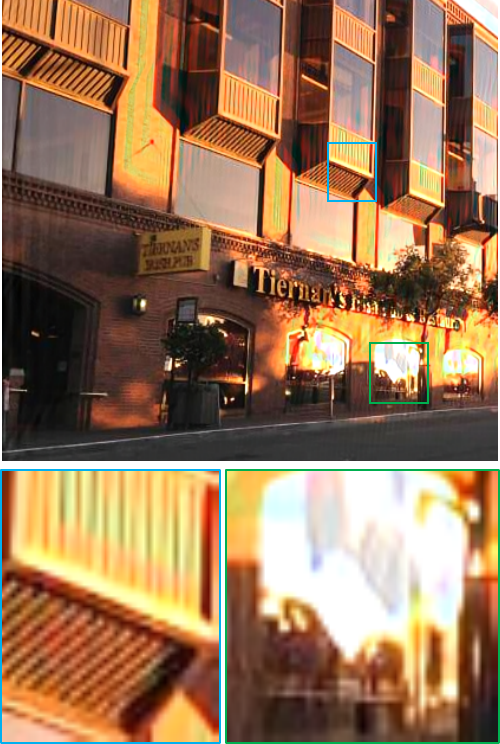}\\
(a) Blurry input & (b) EPLL~\cite{ZoranICCV2012} & (c) CSF~\cite{SchmidtCVPR2014}  & (d) LDT~\cite{DongECCV2018} & (e) IRCNN~\cite{ZhangCVPR2017} & (f) FNBD~\cite{SonICCP2017}  \\[1mm]
\includegraphics[width = 0.16\linewidth]{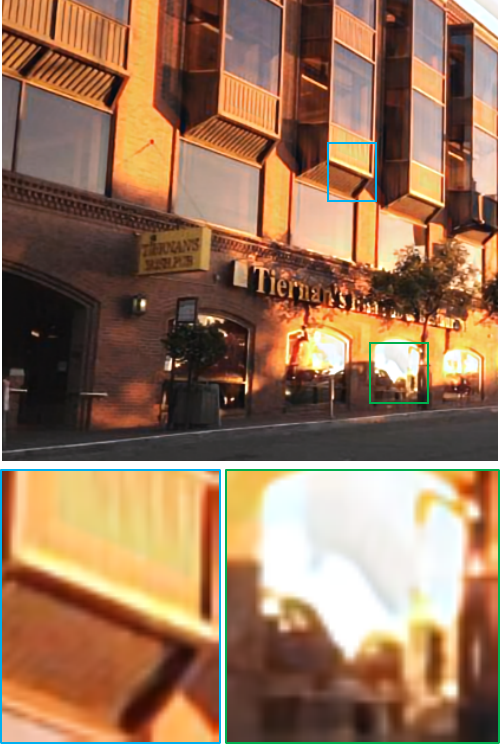}& 
\includegraphics[width = 0.16\linewidth]{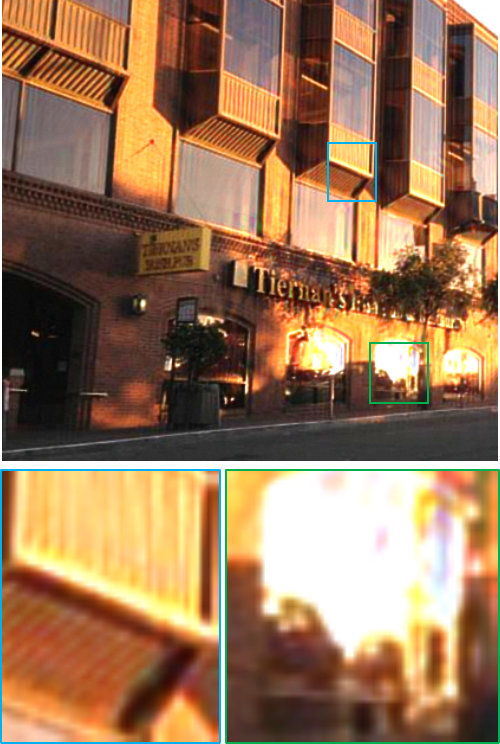}& 
\includegraphics[width = 0.16\linewidth]{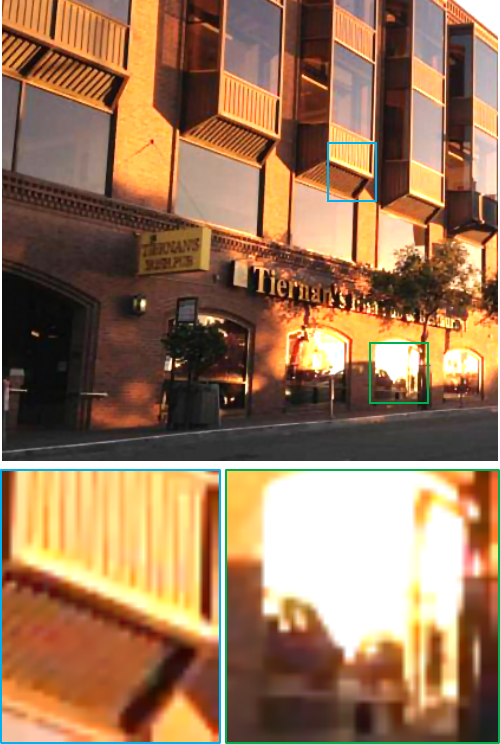}& 
\includegraphics[width = 0.16\linewidth]{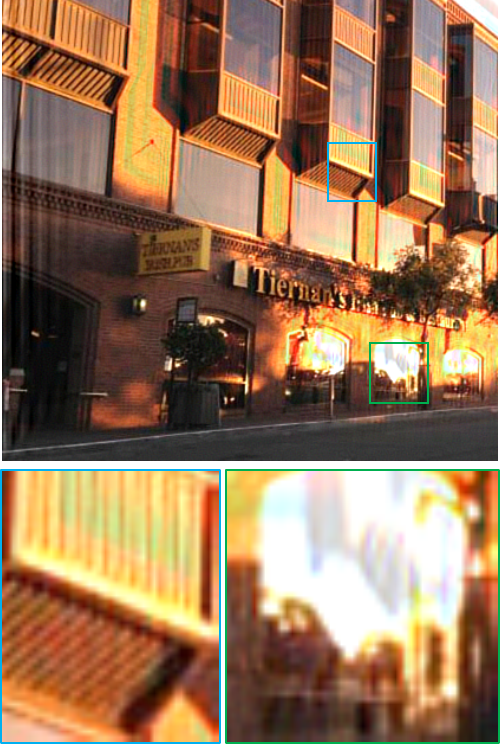}& 
\includegraphics[width = 0.16\linewidth]{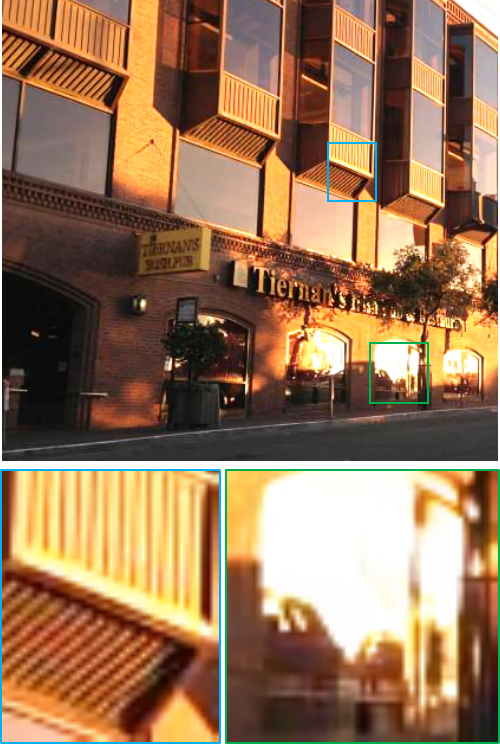} & 
\includegraphics[width = 0.16\linewidth]{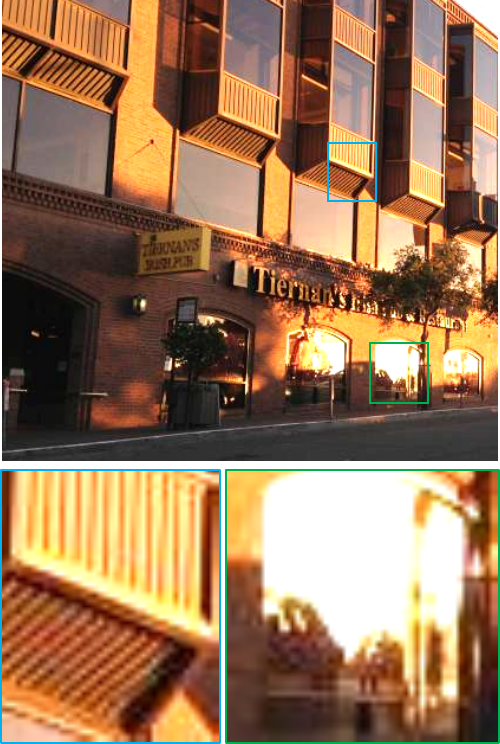}\\
(g) RGDN~\cite{GongTNNLS2020}  & (h) Whyte~\cite{WhyteIJCV2014} & (i) Cho~\cite{ChoICCV2011} & (j) Wiener~\cite{WienerMIT1949} & (k) DWDN+ & (l) Ground truth\\
\end{tabular}
\caption{Example with simulated blur and saturated pixels from~\cite{XuNIPS2014}. The methods \cite{ZoranICCV2012,SchmidtCVPR2014,DongECCV2018,ZhangCVPR2017,SonICCP2017,GongTNNLS2020,WienerMIT1949} do not generate clear images, which contain significant artifacts in \emph{(b)--(g)} and \emph{(j)}. The methods \cite{WhyteIJCV2014,ChoICCV2011} are able to handle saturated pixels, but are not effective at recovering fine detail as shown in the green boxes of \emph{(h)} and \emph{(i)}. In contrast, our approach can preserve small-scale structures and detail as shown in \emph{(k)}.
}
\label{fig:visual_results_sat}
\vspace{-2mm}
\end{figure*}

We compare our approach against several state-of-the-art methods including DMPHN~\cite{ZhangCVPR2019}, EPLL~\cite{ZoranICCV2012}, MLP~\cite{SchulerCVPR2013}, CSF~\cite{SchmidtCVPR2014}, LDT~\cite{DongECCV2018}, FCN~\cite{ZhangJiaweiCVPR2017}, IRCNN~\cite{ZhangCVPR2017}, FDN~\cite{KruseICCV2017}, FNBD~\cite{SonICCP2017}, CPCR~\cite{EboliECCV2020}, and RGDN~\cite{GongTNNLS2020}.
For fair comparison, we retrain the methods~\cite{DongECCV2018,SchmidtCVPR2014,ZhangCVPR2019} with the same training dataset as ours and use the implementation of other methods available online, tuning their adjustable parameters for different datasets and noise levels for best possible results.

\myparagraph{Evaluation results.} We first examine our approach on the dataset of~\cite{LevinCVPR2009} in~\cref{tab:levin_dataset}.
Without manual parameter tuning, our DWDN method outperforms competing approaches by a wide margin, increasing the PSNR by at least $2.09$dB across the noise levels.
Our improved DWDN+ approach further improves the results by at least $0.31$dB.
FNBD~\cite{SonICCP2017} achieves favorable results among the competing methods.
It uses a Wiener deconvolution for preprocessing, a neural network to remove artifacts, and a postprocessing step to recover detail.
In contrast, we explore the Wiener deconvolution in a deep feature space and embed it into our end-to-end network, which is clearly more effective.

\begin{table*}[!t]
  \caption{Quantitative comparison to state-of-the-art methods on the datasets of~\cite{MartinICCV2001} with Gaussian noise and JPEG compression.}
  \vspace{-1mm}
  \label{tab:bsds_dataset_jpeg}
  \centering
  \scriptsize
  \begin{tabularx}{\textwidth}{@{\hspace{3.8mm}}c@{\hspace{3.8mm}}Xc@{\hspace{3.8mm}}c@{\hspace{3.8mm}}c@{\hspace{3.8mm}}c@{\hspace{3.8mm}}c@{\hspace{3.8mm}}c@{\hspace{3.8mm}}c@{\hspace{3.8mm}}c@{\hspace{3.8mm}}c@{\hspace{3.8mm}}c@{\hspace{3.8mm}}c@{}}
    \toprule
$q$ & & EPLL~\cite{ZoranICCV2012}  &MLP~\cite{SchulerCVPR2013} &CSF~\cite{SchmidtCVPR2014} &FCN~\cite{ZhangJiaweiCVPR2017} &IRCNN~\cite{ZhangCVPR2017} &FDN~\cite{KruseICCV2017}&FNBD~\cite{SonICCP2017} &RGDN~\cite{GongTNNLS2020} & Cho~\cite{ChoICCV2011} & DWDN & DWDN+\\
    \midrule
    \multirow{2}{*}{$50$}
&PSNR (dB)&26.70  &25.70  &26.50  &26.52  &25.69  &25.88  &21.57  &25.48  &25.52  &27.80  &\bf27.85\\ 
&SSIM     &0.7552 &0.7101 &0.7341 &0.7490 &0.7111 &0.7381 &0.4981 &0.7084 &0.7097 &0.7877 &\bf0.7914\\
    \midrule
    \multirow{2}{*}{$70$}
&PSNR (dB)&27.71  &26.72  &27.26  &27.52  &26.98  &27.19  &22.54  &26.53  &26.80  &28.67  &\bf28.75 \\ 
&SSIM     &0.7903 &0.7524 &0.7679 &0.7829 &0.7664 &0.7877 &0.5498 &0.7596 &0.7644 &0.8208 &\bf0.8246\\
    \midrule
    \multirow{2}{*}{$90$}
&PSNR (dB)&29.12  &28.02  &28.47  &28.86  &28.79  &29.09  &23.34  &27.98  &28.64  &30.09  &\bf30.21\\ 
&SSIM     &0.8262 &0.7945 &0.8089 &0.8178 &0.8209 &0.8410 &0.5670 &0.8035 &0.8205 &0.8631 &\bf0.8669\\ 
    \bottomrule
  \end{tabularx}
\end{table*}

\begin{figure*}[!t]
\scriptsize\sffamily
\centering
\begin{tabular}{@{}c@{\hspace{1mm}}c@{\hspace{1mm}}c@{\hspace{1mm}}c@{\hspace{1mm}}c@{\hspace{1mm}}c@{}}
\includegraphics[width = 0.16\linewidth]{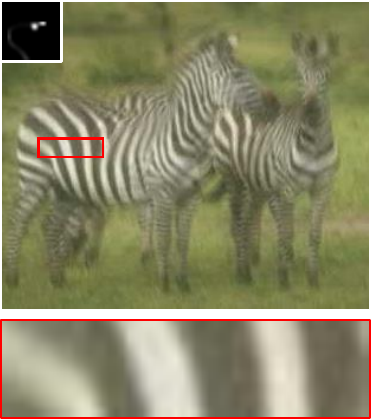}& 
\includegraphics[width = 0.16\linewidth]{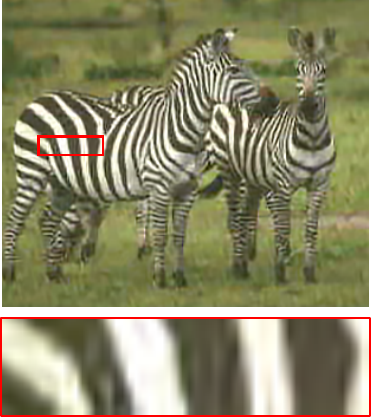}& 
\includegraphics[width = 0.16\linewidth]{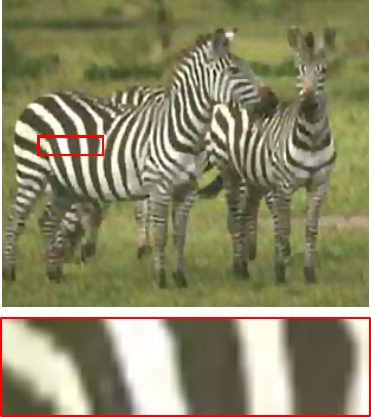}& 
\includegraphics[width = 0.16\linewidth]{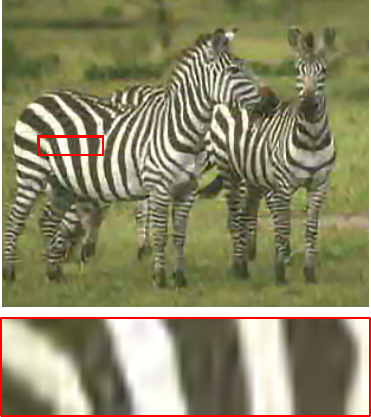}& 
\includegraphics[width = 0.16\linewidth]{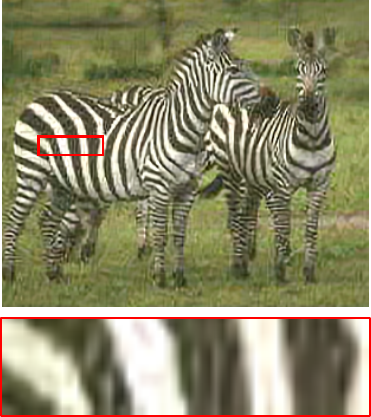} & 
\includegraphics[width = 0.16\linewidth]{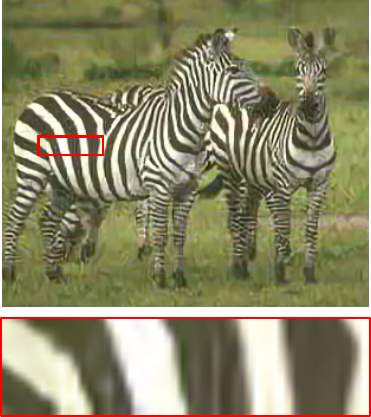}\\
(a) Blurry input & (b) EPLL~\cite{ZoranICCV2012} & (c) CSF~\cite{SchmidtCVPR2014}  & (d) FCN~\cite{ZhangJiaweiCVPR2017} & (e) IRCNN~\cite{ZhangCVPR2017} & (f) FDN \cite{KruseICCV2017}  \\[1mm]
\includegraphics[width = 0.16\linewidth]{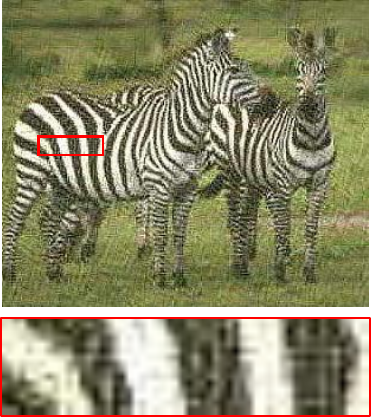}& 
\includegraphics[width = 0.16\linewidth]{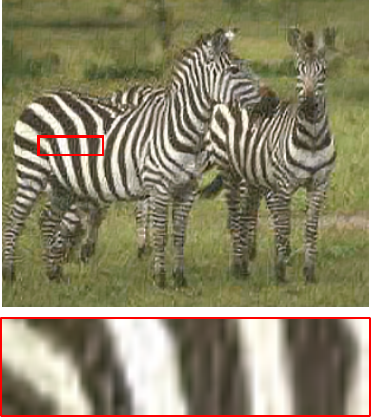}& 
\includegraphics[width = 0.16\linewidth]{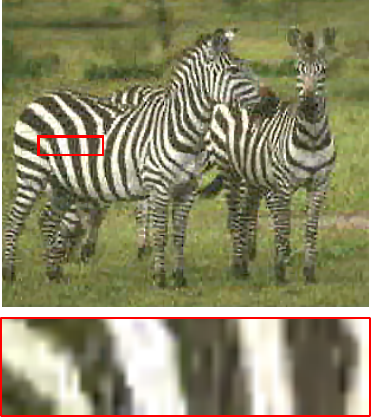}& 
\includegraphics[width = 0.16\linewidth]{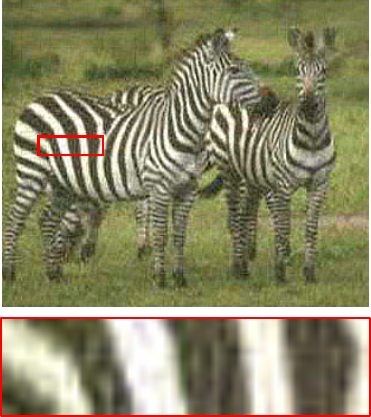}& 
\includegraphics[width = 0.16\linewidth]{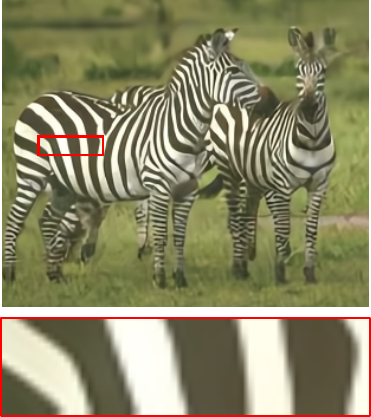} & 
\includegraphics[width = 0.16\linewidth]{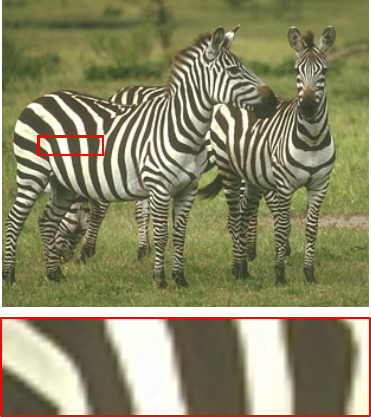}\\
(g) FNBD~\cite{SonICCP2017}  & (h) RGDN~\cite{GongTNNLS2020} & (i) Cho~\cite{ChoICCV2011} & (j) Wiener~\cite{WienerMIT1949} & (k) DWDN+ & (l) Ground truth\\
\end{tabular}
\caption{Example with simulated blur and JPEG compression ($q=50$). The images deblurred by the methods \cite{ZoranICCV2012,ZhangJiaweiCVPR2017,ZhangCVPR2017,KruseICCV2017,SonICCP2017,GongTNNLS2020,ChoICCV2011,WienerMIT1949} contain severe artifacts as shown in \emph{(b)} and \emph{(d)--(j)}. The result obtained by the method \cite{SchmidtCVPR2014} is over-smoothed in \emph{(e)}. Compared to competing methods, our approach can generate a visibly clearer image with sharper small-scale structures and finer detail as shown in \emph{(k)}.
}
\label{fig:visual_results_bsds1jpeg50}
\end{figure*}

We then evaluate the proposed approach on the datasets of~\cite{SunICCP2013} and~\cite{MartinICCV2001} in~\cref{tab:bsds_dataset,tab:sun_dataset} and again observe a significant gain.
The average PSNR of our DWDN method is at least $0.82$dB higher than the best result of~\cite{SonICCP2017} on the dataset of~\cite{MartinICCV2001} and at least $0.48$dB higher than the best competing method on the dataset by Sun \etal~\cite{SunICCP2013}. 
Our enhanced DWDN+ approach further improves the results by at least $0.12$dB on the dataset of~\cite{MartinICCV2001} and $0.19$dB on the dataset of \cite{SunICCP2013}.

\Cref{fig:visual_results_levin5} shows a visual comparison.
A state-of-the-art blind deblurring method~\cite{ZhangCVPR2019} based on an end-to-end network cannot recover a clear image (\cref{fig:visual_results_levin5}(b)), showing the need to further research non-blind methods and backbones.
The results of existing non-blind methods, however, also do not fully recover fine-scale structures or contain severe artifacts (\cref{fig:visual_results_levin5}(c)-(i)).
In contrast, our deep Wiener deconvolution yields a notably clearer image with finer detail and fewer artifacts (\cref{fig:visual_results_levin5}(j)-(k)).
As discussed in~\cref{sec:Analysis_and_Discussion}, our improved results mainly stem from the domain-specific network for image deblurring, which explicitly embeds a feature-based Wiener deconvolution into a deep network.
More visual comparisons are included in the supplemental material.

\subsection{Results with simulated blur and saturated pixels}
\label{ssec:Results with simulated blur and saturated pixels}
\myparagraph{Training dataset.} 
To train the model for blurry images with saturated pixels, we collect 500 low-light images from Flickr.
To synthesize saturated areas, similar to \cite{PanCVPR2016}, we first enlarge the intensity range of the clear images by a factor ranging from \num{1.2} to \num{2}.
We convolve each clear image with a simulated blur kernel \cite{SchmidtPAMI2016} to which $0$--$5\%$ Gaussian noise are added.
Finally, we clip both clear and blurry images to the intensity range of $[0,1]$.

\myparagraph{Test dataset.} 
To evaluate our approach on blurry images with saturated pixels, we collect \num{44} low-light images from the literature \cite{ChoICCV2011,PanCVPR2016,DongICCV2017,XuNIPS2014}. 
Then, we enlarge the intensity of the clear images by a factor of \num{1.2}, which are then convolved with a simulated blur kernel \cite{SchmidtPAMI2016} and added with $0$--$1\%$ Gaussian noise. Finally, both clear and blurry images are clipped to the intensity range of $[0,1]$.
Test and training datasets do not overlap.

\myparagraph{Evaluation results.} 
We quantitatively evaluate our approach on blurry images with saturated pixels in \cref{tab:sat_dataset}, where our DWDN method outperforms other competing approaches by at least \num{0.68}dB and our enhanced DWDN+ approach further improves the image quality by \num{0.44}dB.
As expected, previous methods not specific for handling saturation do not achieve good results.
Whyte \etal\ \cite{WhyteIJCV2014} address the problem of deblurring images with saturated pixels by locating the error-prone bright pixels and estimating them separately with a modified Richardson-Lucy method to reduce ringing.
The method of \cite{ChoICCV2011} employs the EM algorithm to iteratively classify inliers as well as outliers and recover the latent clear image with detected inliers. 
These methods perform well among competing methods, but are not effective at preserving fine-scale structures.
We show visual comparisons in \cref{fig:visual_results_sat}, where the results obtained by \cite{WhyteIJCV2014,ChoICCV2011} are over-smoothed.
In contrast, our DWDN+ approach generates a visibly clearer image in \cref{fig:visual_results_sat}(k).
We also show the result obtained by the classical Wiener deconvolution \cite{WhyteIJCV2014} in \cref{fig:visual_results_sat}(j), which contains severe artifacts.
However, by incorporating our feature-based Wiener deconvolution with the multi-scale feature refinement, our approach can not only avoid introducing artifacts but also recover finer detail as shown in \cref{fig:visual_results_sat}(k).

\subsection{Results with simulated blur and JPEG compression}
\label{ssec:Results with simulated blur and JPEG compression}
\myparagraph{Training dataset.} 
To train the model for handling blurry images with JPEG compression, we use the same training dataset and method as in \cref{ssec:Results with simulated blur and Gaussian noise} to generate blurry image patches.
We add $1\%$ Gaussian noise to each blurry image, which is then distorted by JPEG blocking artifacts.
Different compression quality parameters $q\in\{50, 60, 70, 80, 90\}$ are considered for the JPEG encoder.

\myparagraph{Test dataset.} 
We use the dataset of \cite{MartinICCV2001} to evaluate our approach on blurry images with JPEG compression (no overlap with the training set).
The blurry images are distorted with $1\%$ Gaussian noise and then JPEG compressed with quality parameters of $70, 80,$ and $90$, respectively.

\begin{figure*}[!t]
\scriptsize\sffamily
\centering
\begin{tabular}{@{}c@{\hspace{1mm}}c@{\hspace{1mm}}c@{\hspace{1mm}}c@{\hspace{1mm}}c@{\hspace{1mm}}c@{\hspace{1mm}}c@{}}
\includegraphics[width = 0.16\linewidth]{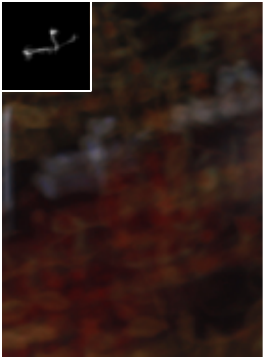}& 
\includegraphics[width = 0.16\linewidth]{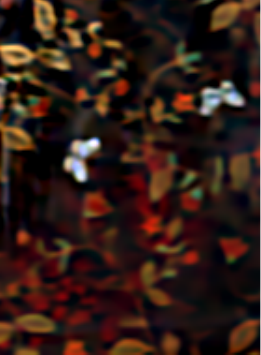}& 
\includegraphics[width = 0.16\linewidth]{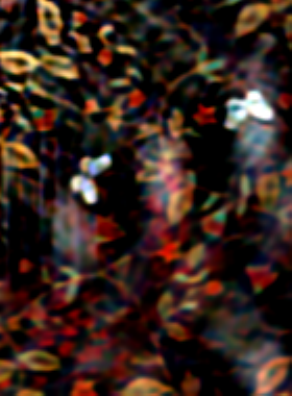}& 
\includegraphics[width = 0.16\linewidth]{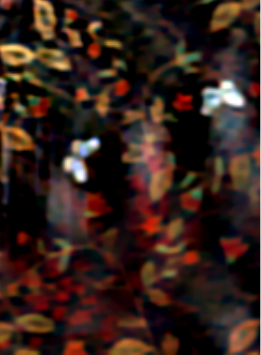}& 
\includegraphics[width = 0.16\linewidth]{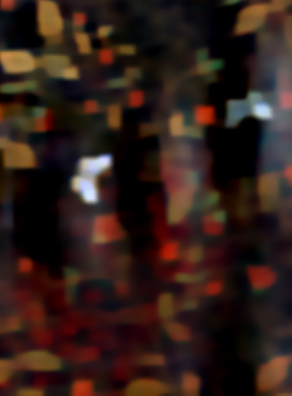}& 
\includegraphics[width = 0.16\linewidth]{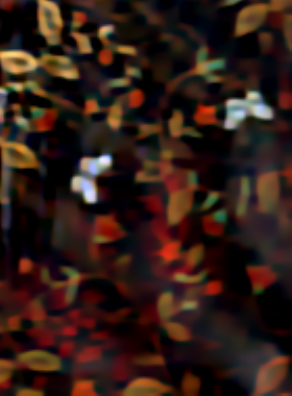} \\
(a) Blurry input & (b) EPLL~\cite{ZoranICCV2012} & (c) MLP \cite{SchulerCVPR2013} & (d) CSF~\cite{SchmidtCVPR2014} & (e) LDT~\cite{DongECCV2018} & (f) FCN \cite{ZhangJiaweiCVPR2017} \\
\includegraphics[width = 0.16\linewidth]{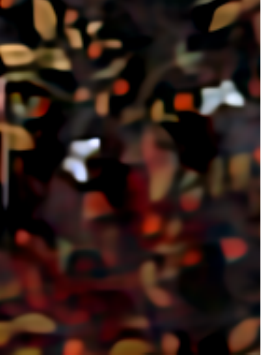}& 
\includegraphics[width = 0.16\linewidth]{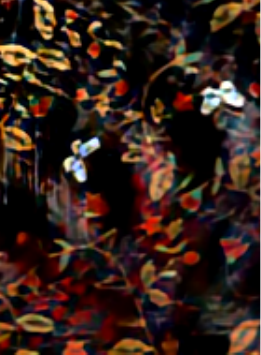}& 
\includegraphics[width = 0.16\linewidth]{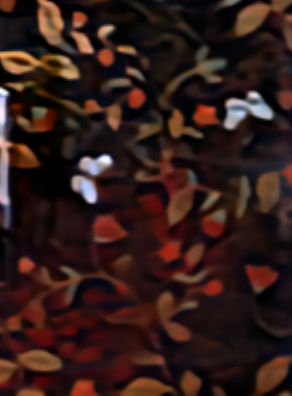}& 
\includegraphics[width = 0.16\linewidth]{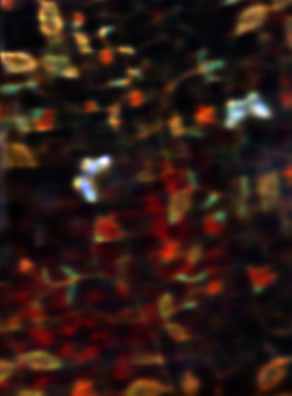}& 
\includegraphics[width = 0.16\linewidth]{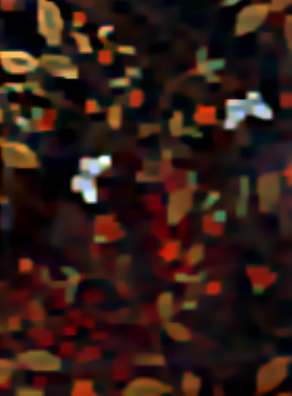}& 
\includegraphics[width = 0.16\linewidth]{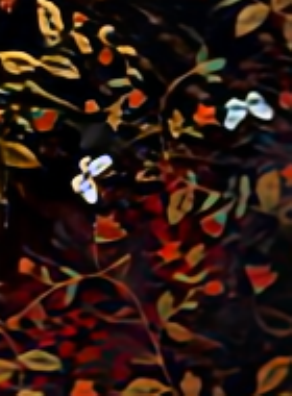} \\
(g) IRCNN~\cite{ZhangCVPR2017} & (h) FNBD~\cite{SonICCP2017} & (i) RGDN~\cite{GongTNNLS2020} & (j) Whyte \cite{WhyteIJCV2014} & (k) Cho \cite{ChoICCV2011} & (l) DWDN+ \\
\end{tabular}
\caption{
Example with real camera shake from~\cite{PanCVPR2016}. Compared with the results in \emph{(b)--(k)}, the deblurred image \emph{(f)} generated by our approach is notably clearer with finer detail and fewer artifacts.
}
\label{fig:visual_results_real1}
\end{figure*}

\myparagraph{Evaluation results.} 
\Cref{tab:bsds_dataset_jpeg} shows that our approach (with a single model) consistently outperforms other competing methods for various JPEG compression quality parameters.
\Cref{fig:visual_results_bsds1jpeg50} shows visual comparisons.
The competing methods \cite{ZoranICCV2012,ZhangJiaweiCVPR2017,ZhangCVPR2017,KruseICCV2017,SonICCP2017,GongTNNLS2020,ChoICCV2011,WienerMIT1949} yield a deblurred result with varying degrees of artifacts as shown in \cref{fig:visual_results_bsds1jpeg50}(b) and (d)--(j).
The method of \cite{SchmidtCVPR2014} does not effectively preserve finer-sale detail in \cref{fig:visual_results_bsds1jpeg50}(c).
In contrast, as shown \cref{fig:visual_results_bsds1jpeg50}(k), our approach recovers much sharper structures, \eg, the texture of the zebra as enclosed in the red box.

\subsection{Results with real blur}
\label{ssec:Results with real blur}
We further evaluate our method on more challenging images with real camera shake, which are more complex and contain unknown types of noise or outliers. 
One example from~\cite{PanCVPR2016} is shown in~\cref{fig:visual_results_real1}, where the blur kernel is estimated by the blind image deblurring method~\cite{DongICCV2017}.
The results recovered by~\cite{SchulerCVPR2013,SchmidtCVPR2014,SonICCP2017,DongECCV2018} exibit visible artifacts, 
\cf~\cref{fig:visual_results_real1}(c)-(e) and (h).
The methods~\cite{ZhangCVPR2017,ZhangJiaweiCVPR2017,ZoranICCV2012,GongTNNLS2020,WhyteIJCV2014,ChoICCV2011} over-smooth the detail in the deblurred images, see~\cref{fig:visual_results_real1}(b), (f)-(g), and (i)-(k).
In contrast, our deep Wiener deconvolution network outputs a visibly clearer image, \cf~\cref{fig:visual_results_real1}(l), where the fine-scale structures of the butterflies and leaves are recovered much better.
More comparisons are included in the supplemental material.

\section{Analysis and Discussion}
\label{sec:Analysis_and_Discussion}
In this section, we first analyze the advantages of the feature-based Wiener deconvolution (\cref{ssec:Effect of the feature-based Wiener deconvolution}), the effect of the multi-scale feature refinement (\cref{ssec:Effect of the feature refinement}), and the robustness of the proposed approach (\cref{ssec:Robustness of the proposed DWDN approach}).
Then we discuss the extension of our method to non-uniform image deblurring in \cref{ssec:Extension to non-uniform image deblurring} and the run-time performance in \cref{ssec:Run-time}. 
%

\begin{table*}[!t]
\scriptsize
\caption{
Effectiveness of the feature-based Wiener deconvolution. All methods are evaluated on the datasets of~\cite{LevinCVPR2009} with $1\%$ Gaussian noise, where the kernel size ranges from $13\times13$ to $35\times35$ pixels. The basic reconstruction network that contains three residual blocks followed by one convolutional layer is denoted as \emph{Basic reconstruction}. The feature refinement networks in \cref{fig: pipeline,fig: pipeline2} are denoted as \emph{Multi-scale refinement} and \emph{Cascaded refinement}, respectively. The baseline method that follows the classical Wiener deconvolution with the Basic reconstruction network is denoted as \emph{Wiener$_{I}$}. The baseline methods that combine the classical Wiener deconvolution with the image gradients, the concatenation of both the input and the image gradients, 9 learned deep features, and 16 learned deep features are denoted as \emph{Wiener$_{G}$}, \emph{Wiener$_{I+G}$}, \emph{Wiener$_{\text{9 features}}$}, and \emph{Wiener$_{\text{16 features}}$}, respectively.
The baseline method that removes the Wiener deconvolution from our proposed network is denoted as \emph{DWDN$_{\text{w/o Wiener}}$}. The baseline methods that replace the learned deep features in our network with the input image, the image gradients, and the concatenation of both the input and the image gradients are denoted as \emph{DWDN$_{I}$}, and \emph{DWDN$_{G}$}, \emph{DWDN$_{I+G}$}.}
  \vspace{-1mm}
  \label{tab:ablation_feature_deconvolution}
  \centering
  \begin{tabularx}{\linewidth}{@{}Xc@{\hspace{2.0mm}}c@{\hspace{2.0mm}}ccc@{\hspace{2.0mm}}c@{\hspace{2.0mm}}cc@{}}
    \toprule
~ &\multicolumn{3}{c}{Feature extraction} & \multirow{2}{*}[-3pt]{\begin{tabular}{@{}c@{}}Wiener\\ deconvolution\end{tabular}} & \multicolumn{3}{c}{Refinement / Reconstruction} & \multirow{2}{*}[-3pt]{\begin{tabular}{@{}c@{}}PSNR (dB)/SSIM\end{tabular}}\\
\cmidrule(lr){2-4}\cmidrule(lr){6-8}
~ & Intensity & Gradients & Deep features & & Basic reconstruction & Multi-scale refinement & Cascaded refinement& \\
\midrule
Wiener~\cite{WienerMIT1949} & \CheckmarkBold & ~ & ~ & \CheckmarkBold & ~ & ~ & ~ & 27.48/0.6981 \\
Wiener$_{I}$ & \CheckmarkBold & ~ & ~ & \CheckmarkBold & \CheckmarkBold & ~ & ~ & 30.32/0.8931 \\
Wiener$_{G}$ & ~ & \CheckmarkBold & ~ & \CheckmarkBold & \CheckmarkBold & ~ & ~ & 18.03/0.6456 \\
Wiener$_{I+G}$ & \CheckmarkBold & \CheckmarkBold & ~ & \CheckmarkBold & \CheckmarkBold & ~ & ~ & 33.37/0.9239 \\
Wiener$_{\text{9 features}}$ & ~ & ~ & \CheckmarkBold & \CheckmarkBold & \CheckmarkBold & ~ & ~ & 34.09/0.9394 \\
Wiener$_{\text{16 features}}$ & ~ & ~ & \CheckmarkBold & \CheckmarkBold & \CheckmarkBold & ~ & ~ & 34.78/0.9478 \\
\midrule
DWDN$_{\text{w/o Wiener}}$ & ~ & ~ & \CheckmarkBold & ~ & ~ & \CheckmarkBold          & ~ & 25.70/0.7893 \\
\midrule
DWDN$_I$ & \CheckmarkBold & ~ & ~ & \CheckmarkBold & ~ & \CheckmarkBold & ~ & 35.13/0.9438 \\
DWDN$_G$ & ~ & \CheckmarkBold & ~ & \CheckmarkBold & ~ & \CheckmarkBold & ~ & 28.79/0.9031 \\ %
DWDN$_{I+G}$ & \CheckmarkBold & \CheckmarkBold & ~ & \CheckmarkBold & ~ & \CheckmarkBold & ~ & 36.52/0.9583 \\ %
DWDN & ~ & ~ & \CheckmarkBold & \CheckmarkBold & ~ & \CheckmarkBold & ~ & 36.90/0.9614 \\ 
\midrule
DWDN+ & ~ & ~ & \CheckmarkBold & \CheckmarkBold & ~ & \CheckmarkBold & \CheckmarkBold & \bf{37.27/0.9634} \\ 
    \bottomrule
  \end{tabularx}
\end{table*}

\subsection{Effect of the feature-based Wiener deconvolution}
\label{ssec:Effect of the feature-based Wiener deconvolution}
Instead of implementing the deconvolution process in the standard image space, we propose a feature-based Wiener deconvolution to better constrain the deconvolution process with useful feature information from the blurry input.
To demonstrate its effectiveness, we first analyze the influence of the feature choice on the classical Wiener deconvolution as well as our proposed deep model. Finally, we visualize and discuss in more detail the learned deep features.

\myparagraph{Influence of the feature choice on classical Wiener deconvolution.}
We first compare the classical Wiener deconvolution \cite{WienerMIT1949} with several baseline methods that combine the classical Wiener deconvolution with various hand-crafted features:
only the image gradients along the vertical and horizontal directions (\emph{Wiener$_G$} for short) and the concatenation of both the blurry image and the image gradients (\emph{Wiener$_{\text{I+G}}$} for short).
To reconstruct the final clear image from the deconvolved features for these baseline methods, we use a basic reconstruction network of three residual blocks followed by one convolutional layer (\ie~the inverse of the feature extraction network, \emph{Basic reconstruction} for short) to go back to the image space.
We then use the dataset~\cite{LevinCVPR2009} ($1\%$ noise level) as described in~\cref{ssec:Results with simulated blur and Gaussian noise} for evaluation.
\Cref{tab:ablation_feature_deconvolution} shows that using the gradient information alone (\ie, Wiener$_{\text{G}}$) is not sufficient, but is effective at improving the deconvolution results when combined with the intensity information from the blurry image, \cf~Wiener and Wiener$_{\text{I+G}}$, increasing the PSNR from \num{27.48}dB to \num{33.37}dB.
For fair comparison, we further compare with a baseline method that follows the classical Wiener deconvolution in image space with the same reconstruction network (\emph{Wiener$_I$} for short).
The comparison of Wiener$_{\text{I}}$ and Wiener$_{\text{I+G}}$ in \cref{tab:ablation_feature_deconvolution} demonstrates that combining useful feature information is able to improve the results of the classical Wiener deconvolution, increasing the PSNR from \num{30.32}dB to \num{33.37}dB.

We then explore the effect of deep features on the classical Wiener deconvolution~\cite{WienerMIT1949}.
As the combination of the blurry image and the image gradients yields \num{9} feature channels, we compare with a baseline that combines the classical Wiener deconvolution with \num{9} deep features learned by the feature extraction network (\emph{Wiener$_{\text{9 features}}$} for short).
We also train a model that learns \num{16} deep features (\emph{Wiener$_{\text{16 features}}$}\footnote{Wiener$_{\text{16 features}}$ is the same as Wiener$_{\text{D}}$ in our conference paper \cite{DongNeurIPS2020}.} for short).
\Cref{tab:ablation_feature_deconvolution} shows that the baseline methods with the learned deep features performs better than the other baseline methods with manually extracted features, \cf~Wiener$_{\text{I+G}}$ and Wiener$_{\text{9 features}}$, increasing the PSNR by \num{0.72}dB to \num{34.09}dB and learning more, useful deep features (\ie, Wiener$_{\text{16 features}}$) can further improve the PSNR to \num{34.78}dB.
We empirically select \num{16} features as a trade-off between image quality and efficiency.
\Cref{tab:ablation_feature_deconvolution} demonstrates that the deep features are more effective at extracting useful information for better deblurring and do not require a manual feature combination compared to fixed feature extractors.

\myparagraph{Influence of the feature choice on the proposed DWDN.}
Our proposed DWDN follows the feature-based Wiener deconvolution with a multi-scale feature refinement (\cref{fig: pipeline2}).
We first compare with a baseline that combines the classical Wiener deconvolution with the multi-scale feature refinement (\emph{DWDN$_{\text{I}}$} for short).
\Cref{tab:ablation_feature_deconvolution} shows that the baseline DWDN$_{\text{I}}$ not only outperforms Wiener$_{\text{I}}$ by a wide margin, but also performs better than Wiener$_{\text{16 features}}$, increasing the PSNR from \num{34.78}dB to \num{35.13}dB.
Thus, the proposed multi-scale feature refinement is able to refine and improve the deconvolved latent result from the deconvolution process.

Next, we discuss whether exploring more feature information is still beneficial in our proposed DWDN model. 
We compare with the baselines that respectively replace the input image in DWDN$_{\text{I}}$ with various hand-crafted or deep features: the image gradients along the vertical and horizontal directions (\emph{DWDN$_G$} for short), the concatenation of both the blurry image and the image gradients (\emph{DWDN$_{I+G}$} for short) and \num{16} learned deep features (\ie, our proposed DWDN method).
The results in~\cref{tab:ablation_feature_deconvolution} demonstrate that our learned deep features still outperform the manual feature combination by about $0.4$dB, which demonstrates that the feature extraction network can effectively learn useful features for the embedded Wiener deconvolution step to facilitate better image deblurring.
In addition, finer-scale detail is better modeled in the feature space. Particularly, the deep feature extractor in our end-to-end network can adaptively learn useful features for high-quality image deblurring (\cref{fig:motivation}).

\begin{figure}[t]
\scriptsize\sffamily
\centering
\begin{tabular}{@{}c@{\hspace{1mm}}c@{\hspace{1mm}}c@{\hspace{1mm}}c@{\hspace{1mm}}c@{\hspace{1mm}}c@{\hspace{1mm}}c@{\hspace{1mm}}c@{}}
Blurry input & \multicolumn{5}{c}{Features $\{\textbf{F}_i \textbf{y}\}$ learned by our feature extraction network}\\
\includegraphics[width = 0.155\linewidth]{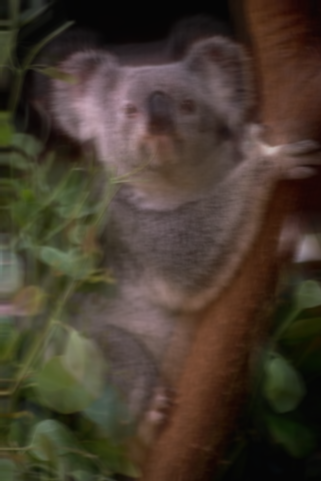}&
\includegraphics[width = 0.155\linewidth]{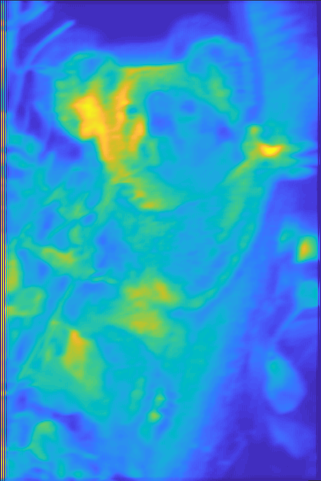}& 
\includegraphics[width = 0.155\linewidth]{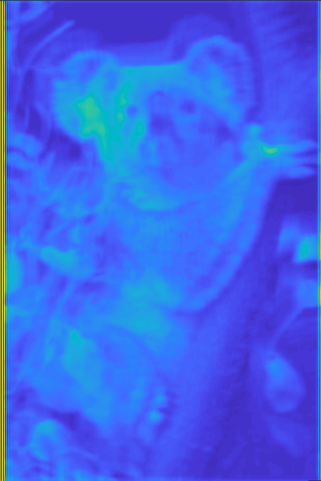}& 
\includegraphics[width = 0.155\linewidth]{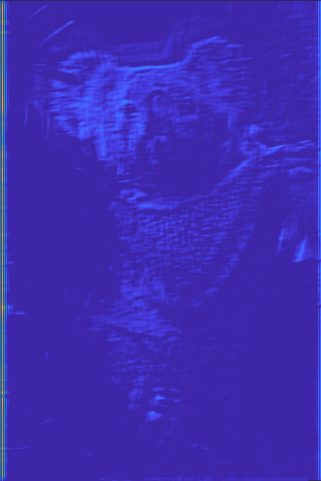}& 
\includegraphics[width = 0.155\linewidth]{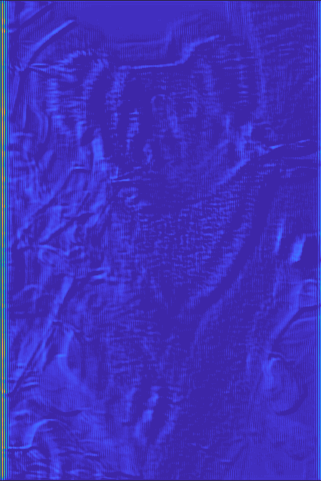}&
\includegraphics[width = 0.155\linewidth]{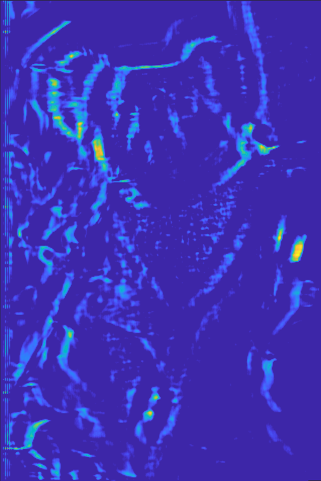}\\[-0.5mm]
\multicolumn{6}{c}{(a) Results on the blurry input without noise} \\[0.75mm]
\includegraphics[width = 0.155\linewidth]{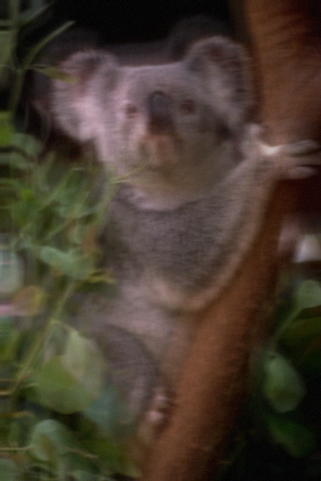}&
\includegraphics[width = 0.155\linewidth]{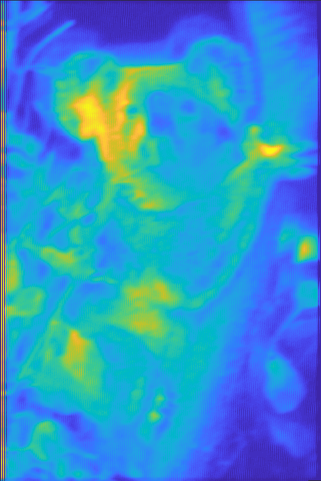}& 
\includegraphics[width = 0.155\linewidth]{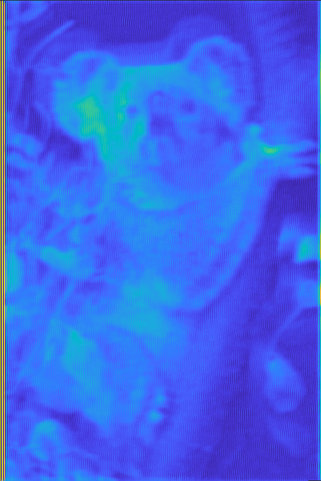}& 
\includegraphics[width = 0.155\linewidth]{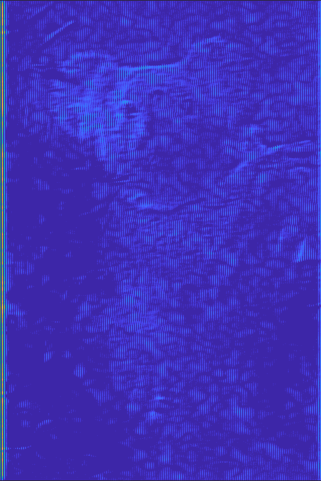}& 
\includegraphics[width = 0.155\linewidth]{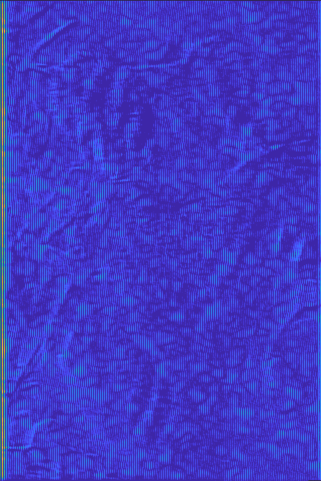}&
\includegraphics[width = 0.155\linewidth]{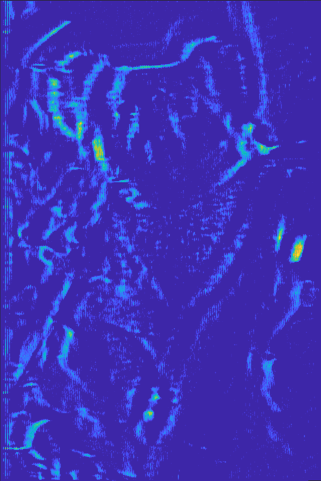}\\[-0.5mm]
\multicolumn{6}{c}{(b) Results on the blurry input with $1\%$ Gaussian noise} \\[0.75mm]
\includegraphics[width = 0.155\linewidth]{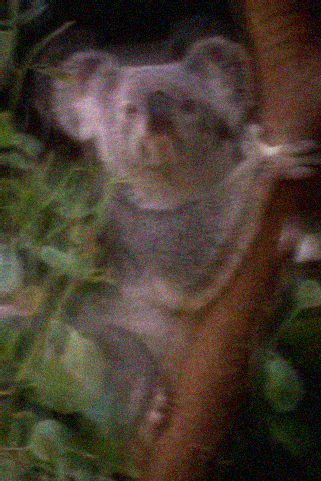}&
\includegraphics[width = 0.155\linewidth]{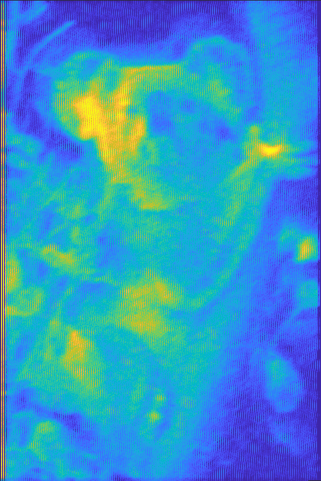}& 
\includegraphics[width = 0.155\linewidth]{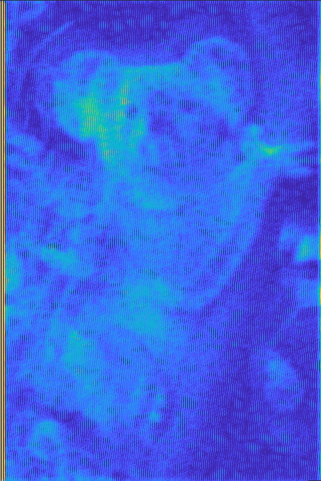}& 
\includegraphics[width = 0.155\linewidth]{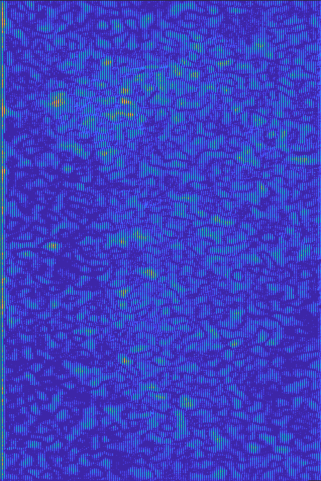}& 
\includegraphics[width = 0.155\linewidth]{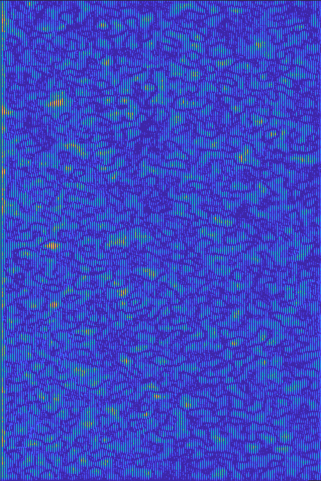}&
\includegraphics[width = 0.155\linewidth]{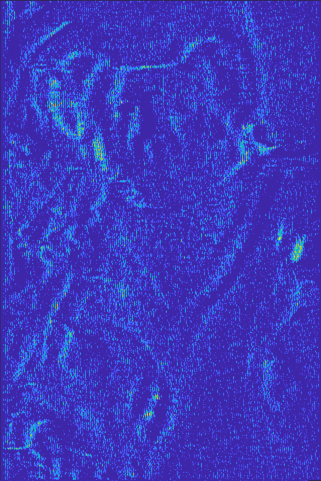}\\[-0.5mm]
\multicolumn{6}{c}{(c) Results on the blurry input with $5\%$ Gaussian noise} \\[0.75mm]
\includegraphics[width = 0.155\linewidth]{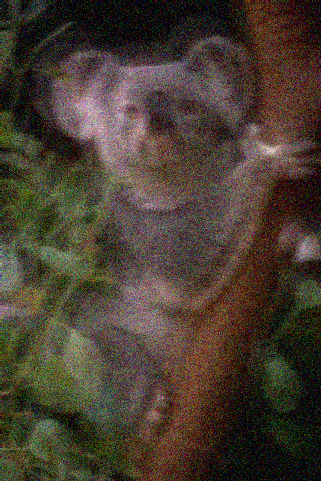}&
\includegraphics[width = 0.155\linewidth]{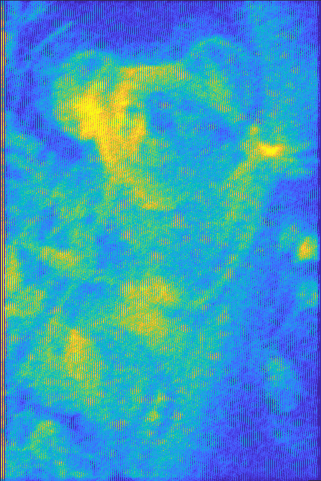}& 
\includegraphics[width = 0.155\linewidth]{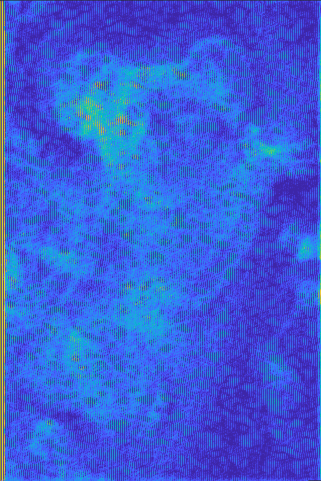}& 
\includegraphics[width = 0.155\linewidth]{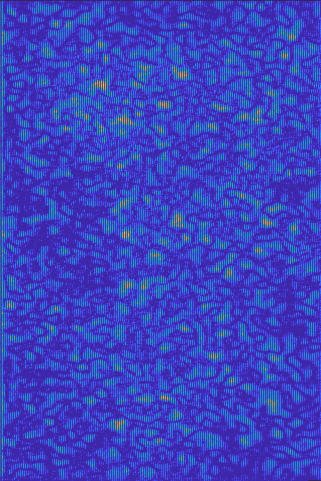}& 
\includegraphics[width = 0.155\linewidth]{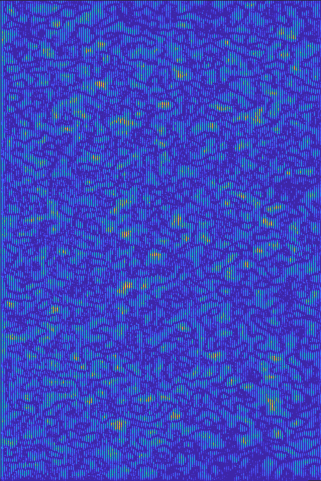}&
\includegraphics[width = 0.155\linewidth]{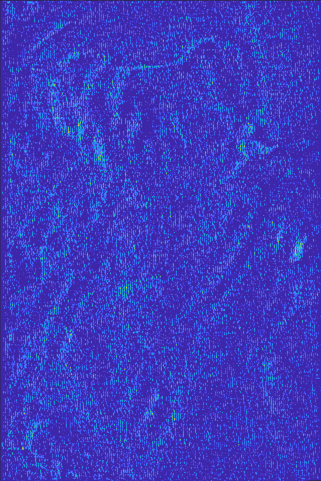}\\[-0.5mm]
\multicolumn{6}{c}{(d) Results on the blurry input with $10\%$ Gaussian noise} \\
\end{tabular}
\caption{Learned features $\{\textbf{F}_i \textbf{y}\}$ for blurry images with noise of different levels. From \emph{(a)} to \emph{(d)}, the noise level in the blurry input becomes higher and the extracted features contain more noise.
}
\label{fig:role_feature_extractor}
\end{figure}

\begin{figure*}[t]
\footnotesize\sffamily
\centering
\begin{tabular}{@{}c@{\hspace{1mm}}c@{\hspace{1mm}}c@{\hspace{1mm}}c@{\hspace{1mm}}c@{\hspace{1mm}}c@{\hspace{1mm}}c@{\hspace{1mm}}c@{}}
\includegraphics[width = 0.12\linewidth,height=2.7cm]{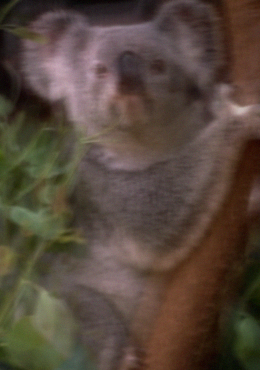}&
\includegraphics[width = 0.12\linewidth,height=2.7cm]{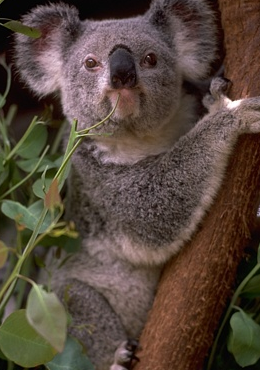}& 
\includegraphics[width = 0.12\linewidth,height=2.7cm]{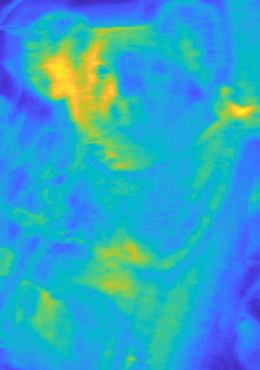}& 
\includegraphics[width = 0.12\linewidth,height=2.7cm]{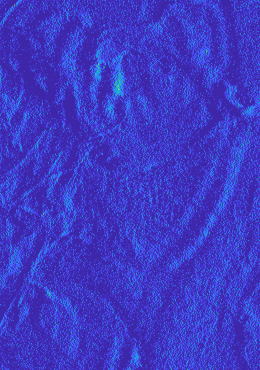}& 
\includegraphics[width = 0.12\linewidth,height=2.7cm]{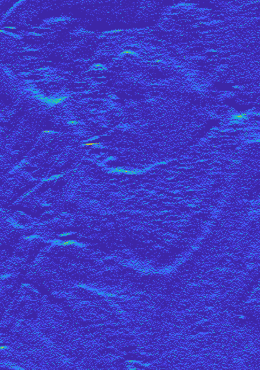}&
\includegraphics[width = 0.12\linewidth,height=2.7cm]{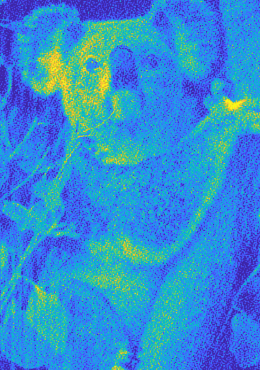}&
\includegraphics[width = 0.12\linewidth,height=2.7cm]{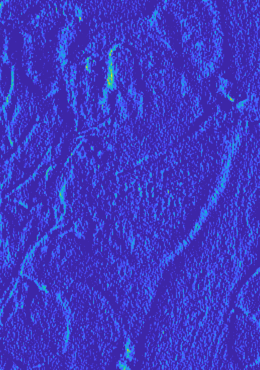}& 
\includegraphics[width = 0.12\linewidth,height=2.7cm]{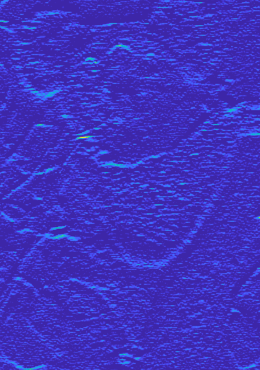}\\[-0.5mm]
(a) & (b)  & (c) & (d) & (e)  & (f) & (g) & (h)  \\[0.75mm]
\includegraphics[width = 0.12\linewidth,height=2.7cm]{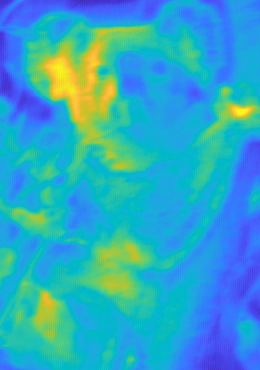}& 
\includegraphics[width = 0.12\linewidth,height=2.7cm]{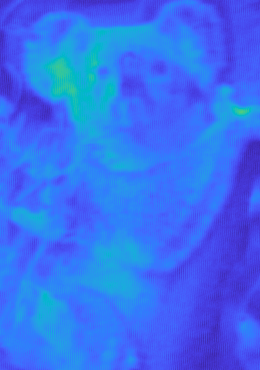}& 
\includegraphics[width = 0.12\linewidth,height=2.7cm]{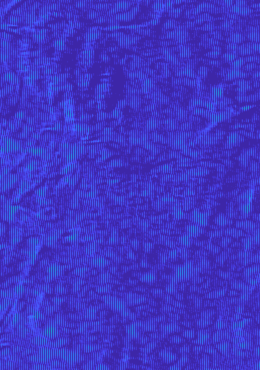}& 
\includegraphics[width = 0.12\linewidth,height=2.7cm]{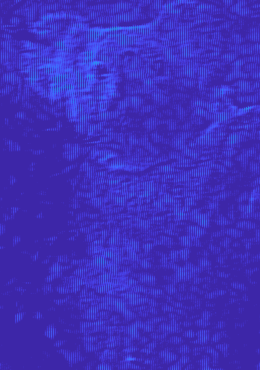}& 
\includegraphics[width = 0.12\linewidth,height=2.7cm]{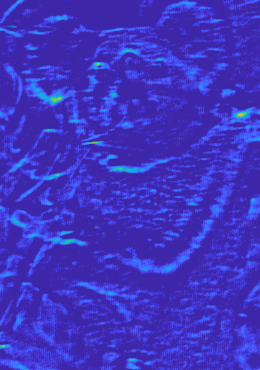}& 
\includegraphics[width = 0.12\linewidth,height=2.7cm]{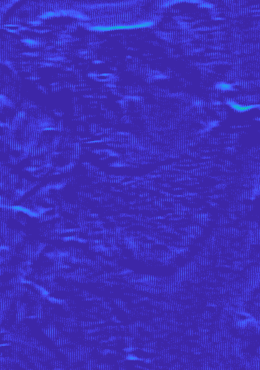}& 
\includegraphics[width = 0.12\linewidth,height=2.7cm]{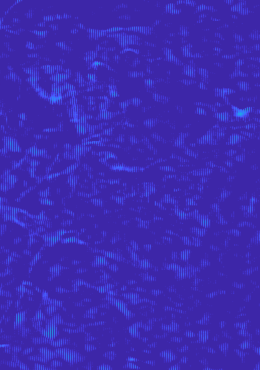}& 
\includegraphics[width = 0.12\linewidth,height=2.7cm]{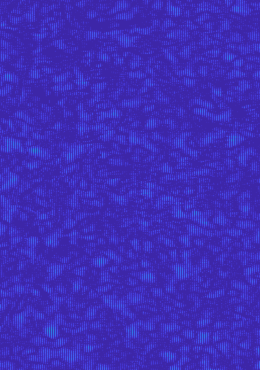}\\[-0.5mm]
\multicolumn{8}{c}{(i)} \\[0.75mm]
\includegraphics[width = 0.12\linewidth,height=2.7cm]{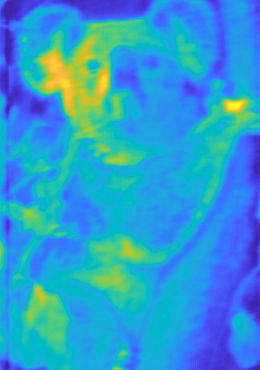}& 
\includegraphics[width = 0.12\linewidth,height=2.7cm]{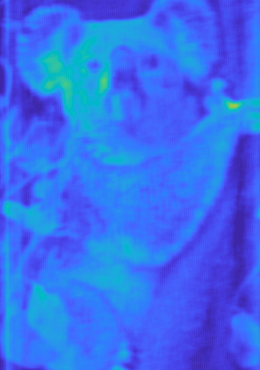}& 
\includegraphics[width = 0.12\linewidth,height=2.7cm]{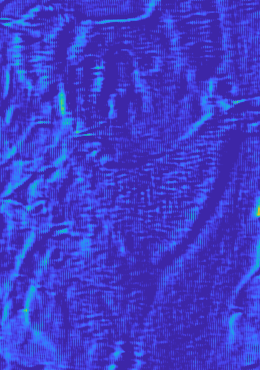}&
\includegraphics[width = 0.12\linewidth,height=2.7cm]{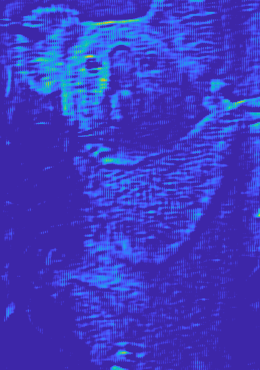}& 
\includegraphics[width = 0.12\linewidth,height=2.7cm]{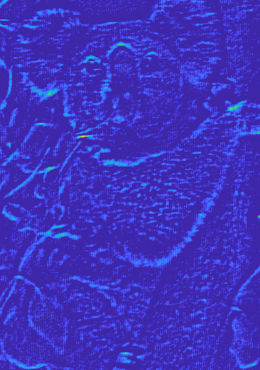}& 
\includegraphics[width = 0.12\linewidth,height=2.7cm]{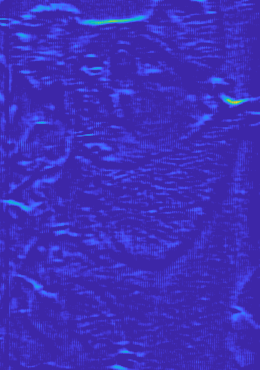}& 
\includegraphics[width = 0.12\linewidth,height=2.7cm]{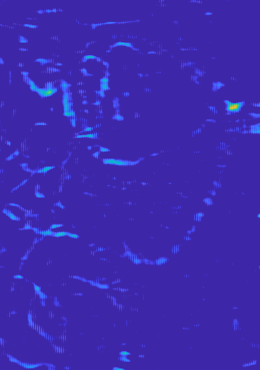}& 
\includegraphics[width = 0.12\linewidth,height=2.7cm]{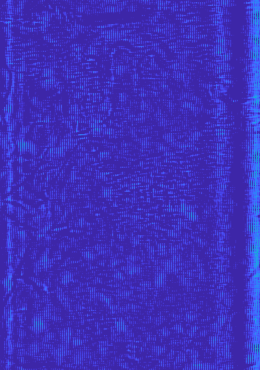}\\[-0.5mm]
\multicolumn{8}{c}{(j)} \\
\end{tabular}
\caption{Illustration of learned deep features. \emph{(a)} Blurry input. \emph{(b)} Ground truth. \emph{(c)--(e)} are the visualization of one channel of the blurry input, the image gradient along the vertical direction, and the image gradient along the horizontal direction, respectively. \emph{(f)--(h)} are the deconvolved results corresponding to \emph{(c)--(e)}. \emph{(i)} and \emph{(j)} visualize some features $\{\textbf{F}_i \textbf{y}\}$ learned by DWDN's feature extraction network and the corresponding deconvolved results $\{\textbf{F}_i \hat{\textbf{x}}\}$, respectively.
}
\label{fig:features}
\end{figure*}

To investigate the role of the Wiener deconvolution in the proposed DWDN, we additionally compare with a baseline that removes the Wiener deconvolution (\emph{DWDN$_{\text{w/o Wiener}}$} for short, which contains the feature extraction network followed by the feature refinement network and thus is not guided by the blur kernel) and train this baseline using the same settings as ours.
\Cref{tab:ablation_feature_deconvolution} shows that the PSNR of DWDN$_{\text{w/o Wiener}}$ drops to \num{25.70}dB, which is the worst one among all baselines. 
The comparisons among Wiener$_{\text{I}}$, DWDN$_{\text{w/o Wiener}}$, and DWDN illustrate the importance of the feature-based Wiener deconvolution on the end-to-end network for non-blind image deblurring, which can effectively incorporate kernel information and leverage deep features.

Also note that our encoder-decoder based multi-scale refinement network is much better at recovering the clear image from the deconvolved features than the basic reconstruction network, \cf~Wiener$_{\text{16 features}}$ and DWDN, increasing the PSNR by $2.05$dB to $36.90$dB.
Strengthening the feature refinement with a cascade architecture (\cref{fig: pipeline}) in DWDN+ can further improve the image quality by increasing the PSNR to $37.27$dB.
Overall, the benefit of our approach stems from the joint effect of feature extraction, Wiener deconvolution, and feature refinement.

\myparagraph{Visualization of learned deep features.}
To intuitively illustrate what the proposed feature extraction network learns, we show some learned features $\{\textbf{F}_i \textbf{y}\}$ in~\cref{fig:features}(i), where the corresponding results deconvolved by \cref{eq:deconvolution,eq:wiener_filter}  (\ie $\{\textbf{F}_i \hat{\textbf{x}}\}$) are shown in~\cref{fig:features}(j).
The blurry input and ground truth are shown in~\cref{fig:features}(a)--(b).
For better understanding, we also show the visualization of one channel of the blurry input and the image gradients along two directions in~\cref{fig:features}(c)--(e).
Their corresponding deconvolved results are shown in~\cref{fig:features}(f)--(h).
Compared to the intensity information and the gradient information in~\cref{fig:features}(c)--(h), the learned deep features in~\cref{fig:features}(i)--(j) contain much richer feature information, facilitating the reconstruction of high-quality images.
In addition, by integrating feature extraction, deep Wiener deconvolution, and feature refinement into an end-to-end network, the feature extraction network can automatically learn useful feature information from the blurry input for better deblurring.
Thus, the proposed method does not require a manual combination of features.

\myparagraph{Role of the feature extraction network.}
As stated in \cref{ssec:feature_deconvolution}, we propose to obtain powerful feature extractors $\{\textbf{F}_i\}$ using deep neural networks that provide more useful information for a subsequent Wiener deconvolution.
However, one may actually wonder whether the feature extraction network acts as a denoiser, leading to the observed robustness of the proposed method to various noise levels.
To answer this question, we further compare the learned features $\{\textbf{F}_i \textbf{y}\}$ for blurry images without and with noise of different noise levels ($1\%, 5\%, 10\%$) in~\cref{fig:role_feature_extractor}.
Each column from the second to sixth columns in ~\cref{fig:role_feature_extractor} exhibits the same feature channel learned by the proposed feature extraction network.
For different noise levels, the learned features display similar noise characteristics as the input, \ie~the higher the noise level of the blurry image is, the more noise the learned feature contains.
Therefore, the feature extraction network does not act as a denoiser, but plays the role of exploiting useful feature information that supports the subsequent feature deconvolution and refinement.

\myparagraph{Linear \emph{vs.} piece-wise linear feature extraction network.}
As stated in \cref{ssec:feature_deconvolution}, the derivations of the feature-based Wiener deconvolution strictly hold in a linear feature space.
Moreover, we are interested to leverage powerful learned feature extractors $\{\textbf{F}_i\}$ beyond hand-crafted ones.
To this end, we develop a feature extraction network to estimate $\{\textbf{F}_i \textbf{y}\}$.
As the feature extraction network with ReLUs is piece-wise linear, the linearity assumption of the Wiener deconvolution holds only locally~\cite{MontufarNIPS2014,LeeICLR2019}.
To evaluate the feasibility and effectiveness of this piece-wise linear feature extraction network, we compare with a baseline that removes the ReLUs in the proposed feature extraction network.
Here we focus on the effect of learned deep features, disable our proposed feature refinement network for all the baseline methods, and use the same basic reconstruction network as in~\cref{tab:ablation_feature_deconvolution} to reconstruct the final image from the deconvolved deep features.
We compare the effect of using features extracted by the proposed network (\ie, Wiener$_{\text{16 features}}$) and the linear feature extraction network (\emph{Wiener$_{\text{16 features}}$ w/ learned linear features} for short) on the classical Wiener deconvolution. \Cref{tab:ablation_deep_features} shows that the method with the features extracted by the proposed piece-wise linear (deep) feature extraction network performs much better than that using only a linear feature extraction network, increasing the PSNR by at least $1.09$dB on the dataset of~\cite{LevinCVPR2009}, $1.00$dB on the dataset of~\cite{MartinICCV2001}, and $1.49$dB on the dataset of~\cite{SunICCP2013}.
Hence, even if the assumptions of the Wiener deconvolution only hold locally for the piece-wise linear deep features, the feature refinement can compensate that and results in a significant benefit from the more powerful feature extractor.

\begin{table}[!t]
\scriptsize
  \caption{Effectiveness of the deep features learned by our piece-wise linear feature extraction network, evaluated on the datasets of~\cite{LevinCVPR2009}, \cite{MartinICCV2001}, and~\cite{SunICCP2013} (PSNR in dB/SSIM). We compare the baseline method that combines the classical Wiener deconvolution with the proposed deep feature extraction network (\emph{Wiener$_{\text{16 features}}$}) with a baseline that uses a linear feature extraction network (\emph{Wiener$_{\text{16 features}}$ w/ learned linear features}).}
  \vspace{-1mm}
  \label{tab:ablation_deep_features}
  \centering
  \begin{tabularx}{\linewidth}{@{}l@{\hspace{2.0mm}}X@{\hspace{2.0mm}}c@{\hspace{2.0mm}}c@{}}
    \toprule
Dataset & Noise level  & Wiener$_{\text{16 features}}$ w/ learned linear features & Wiener$_{\text{16 features}}$\\
    \midrule
\multirow{3}{*}{\cite{LevinCVPR2009}}
&$1\%$                   &33.69/0.9133      &\bf34.78/0.9478 \\
&$3\%$                   &29.80/0.8343      &\bf31.18/0.8942 \\
&$5\%$                   &28.09/0.7888      &\bf29.39/0.8553 \\
\midrule
\multirow{1}{*}{\cite{MartinICCV2001}}
&$1\%$                   &31.69/0.8815      &\bf32.69/0.9127 \\
\midrule
\multirow{1}{*}{\cite{SunICCP2013}}
&$1\%$                   &29.46/0.8472      &\bf30.95/0.8836 \\
\bottomrule
  \end{tabularx}
\end{table}

\begin{table}[!t]\scriptsize
\caption{Effect of the multi-scale configuration (PSNR in dB/SSIM). All methods are evaluated on the dataset of~\cite{LevinCVPR2009} with $1\%$ Gaussian noise, where the kernel size ranges from $13\times13$ to $27\times27$ pixels. We denote the baseline method without the multi-scale refinement as \emph{DWDN$_{\text{w/o multi-scale}}$}.}
\label{tab:ablation_feature_refinement}
\vspace{-1mm}
\centering
\scriptsize
\begin{tabularx}{\linewidth}{@{}lXccc@{}}
    \toprule
Noise level      & Kernel size & 13--19 & 33--39 & 53--59 \\
    \midrule
\multirow{2}{*}{$1\%$}
&DWDN$_{\text{w/o multi-scale}}$ &33.46/0.9232 &29.23/0.8398 &27.45/0.7995\\
&DWDN                            &33.57/0.9254 &29.47/0.8437 &27.73/0.8127\\
    \midrule
\multirow{2}{*}{$3\%$}
&DWDN$_{\text{w/o multi-scale}}$ &29.93/0.8475 &26.71/0.7370 &25.20/0.6918\\
&DWDN                            &30.09/0.8519 &27.00/0.7510 &25.57/0.7080\\
    \bottomrule
  \end{tabularx}
\end{table}

\subsection{Effect of the feature refinement}
\label{ssec:Effect of the feature refinement}
To reconstruct high-quality images from the latent features deconvolved by the feature-based Wiener deconvolution, we develop a multi-scale feature refinement module in a coarse-to-fine manner and employ the cascade architecture at each image scale to better restore the final clear image.
Next, we will discuss the effect of the multi-scale configuration and the cascade architecture used in the proposed feature refinement.

\myparagraph{Effectiveness of the multi-scale configuration.}
Although multi-scale approaches have been widely used in blind image deblurring~\cite{PanPAMI2016,TaoCVPR2018} to facilitate extracting multi-scale structural information, its effect on non-blind deblurring based on deep learning is still unknown.
To demonstrate the effect of the multi-scale refinement in our non-blind deblurring approach, we compare with a baseline method without using the multi-scale refinement (\ie~we set $L=1$ in \cref{eq:multi-scale,eq:loss}, \emph{DWDN$_{\text{w/o multi-scale}}$} for short) that is implemented in the same way as ours otherwise, and evaluate on the dataset of~\cite{MartinICCV2001} with different kernel sizes and noise levels.
\Cref{tab:ablation_feature_refinement} shows that, depending on the kernel size in the ranges $13$--$19$, $33$--$39$, and $53$--$59$ pixels, the PSNR of our DWDN method is $0.16$dB, $0.29$dB, and $0.37$dB higher, respectively (with $3\%$ Gaussian noise), than the baseline method without using multi-scale refinement.
This demonstrates that our multi-scale refinement clearly improves the results, especially in the challenging cases of large blurs (in which many methods including blind ones are known to have difficulties) and strong noise.

\myparagraph{Effectiveness of the cascade architecture.}
Different from our conference paper \cite{DongNeurIPS2020} that adopts a standard encoder-decoder network at each image scale of the feature refinement module, we here propose to employ cascades of encoder-decoders to better refine the deconvolved features and reconstruct the final clear image as mentioned in \cref{ssec:multi-scale_feature_refinement}. 
To analyze the effect of the cascade architecture and the residual learning used in earlier cascade stages, we compare our original DWDN method (without using both cascades and residual learning), a baseline that improves DWDN with \num{2} cascades (\emph{DWDN$_{\text{2 cascades}}$} for short), a baseline method that removes residual learning from our improved DWDN+ (\emph{DWDN+$_{\text{w/o RL}}$} for short),  and our DWDN+ approach in \cref{tab:ablation_cascaded}.
The comparisons show that using more cascade stages at each image scale can clearly improve the image quality by \num{0.22}dB with \num{2} cascades and \num{0.42}dB with \num{3} cascades.
With residual learning, our DWDN+ approach can preserve fine-scale image information better and further improve the final results as shown in \cref{tab:ablation_cascaded}.

\begin{table}[!t]\scriptsize
\caption{Effectiveness of the cascade architecture. All the methods are evaluated on the dataset of~\cite{SunICCP2013} with $1\%$ Gaussian noise. We denote the baseline that improves DWDN with \num{2} cascades as \emph{DWDN$_{\text{2 cascades}}$} and the baseline that removes residual learning from DWDN+ as \emph{DWDN+$_{\text{w/o RL}}$}.}
\label{tab:ablation_cascaded}
\vspace{-1mm}
\begin{tabularx}{\linewidth}{@{}Xc@{\hspace{3.0mm}}c@{\hspace{3.0mm}}c@{\hspace{3.0mm}}c@{}}
    \toprule
~ &\multicolumn{2}{c}{Cascades} &\multirow{2}{*}[-3pt]{Residual learning} &\multirow{2}{*}[-3pt]{PSNR(dB)/SSIM} \\
\cmidrule(lr){2-3}
~ &~~~~~~2~~~~~~ & ~~~~~~3~~~~~~            \\
    \midrule
DWDN                       &\XSolidBrush   &\XSolidBrush   &\XSolidBrush   &34.05/0.9225\\
DWDN$_{\text{2 cascades}}$ &\CheckmarkBold &\XSolidBrush   &\XSolidBrush   &34.27/0.9234\\ %
DWDN+$_{\text{w/o RL}}$    &\XSolidBrush   &\CheckmarkBold &\XSolidBrush   &34.47/0.9280\\ %
DWDN+                      &\XSolidBrush   &\CheckmarkBold &\CheckmarkBold &\bf34.62/0.9296\\
    \bottomrule
  \end{tabularx}
\end{table}

\begin{table}[t]
\scriptsize
  \caption{Robustness of the proposed deep Wiener devoncolution network to various noise levels, evaluated on the dataset of Levin \etal~\cite{LevinCVPR2009} (PSNR in dB/SSIM).
  The instances of our model that are specifically trained with only one Gaussian noise level is denoted as \emph{DWDN$_{\text{specific}}$}.}
  \vspace{-1mm}
  \label{tab:ablation_robustness}
  \centering
  \begin{tabularx}{\linewidth}{@{}Xc@{\hspace{3.0mm}}c@{\hspace{3.0mm}}c@{\hspace{3.0mm}}c@{}}
    \toprule
Noise level   &IRCNN~\cite{ZhangCVPR2017} &FNBD~\cite{SonICCP2017} & DWDN$_{\text{specific}}$  & DWDN \\
    \midrule
$1\%$ &34.33/0.9210 &34.81/0.9398 &\bf37.02/0.9624 &36.90/0.9614 \\
$3\%$ &30.04/0.8156 &30.63/0.8658 &\bf32.88/0.9191 &32.77/0.9179 \\
$5\%$ &28.51/0.7762 &27.93/0.7759 &{\bf30.80}/0.8846 &30.77/\bf0.8857 \\
$10\%$&26.14/0.7078 &23.98/0.5756 &\bf28.04/0.8241 &27.78/0.8189 \\
    \bottomrule
  \end{tabularx}
\end{table}

\subsection{Robustness of the proposed DWDN approach}
\label{ssec:Robustness of the proposed DWDN approach}

\myparagraph{Robustness to differing noise levels.}
As mentioned in \cref{ssec:feature_deconvolution}, $s_i^x$ and $s_i^n$ denote $\mathbb{E} \big( | \textbf{F}_i \textbf{x} |^2 \big)$ and $\mathbb{E} \big( |\textbf{F}_i\textbf{n}|^2 \big)$, respectively.
However, in real applications, it is hard to accurately calculate these expectations.
Similar to existing methods~\cite{SonICCP2017,XuNIPS14}, we estimate $s_i^x$ by the standard deviation of the blurry feature $\textbf{F}_i \textbf{y}$. $s_i^n$ is estimated by the variance of the difference between the blurry feature $\textbf{F}_i \textbf{y}$ and the mean-filtered result of $\textbf{F}_i \textbf{y}$.
Note that $s_i^n$ is adaptively computed from the blurry feature $\textbf{F}_i \textbf{y}$ and related to the noise level in $\textbf{F}_i \textbf{y}$.
Thus, the proposed network is able to handle blurry images with various noise levels.
This is also referred to as being noise-blind \cite{JinCVPR2017}.
To demonstrate the robustness of the proposed DWDN approach to various noise levels, we compare our \emph{single} model (which is trained in images with various Gaussian noise levels, ranging from $0$--$5\%$) with instances of our model that are specifically trained with only one Gaussian noise level of either $1\%$, $3\%$, or $5\%$ (\emph{DWDN$_{\text{specific}}$} for short).
\Cref{tab:ablation_robustness} shows that the proposed noise-blind model obtains similar results to the models with noise-specific training across various noise levels, \eg, $36.90$dB compared to $37.02$dB for $1\%$ Gaussian noise and  $30.77$dB compared to $30.80$dB for $5\%$ Gaussian noise. 
It is thus clearly noise robust.

As mentioned in \cref{ssec:Results with simulated blur and Gaussian noise}, our normal training dataset contains blurry images with Gaussian noise of various noise levels, ranging from $0$ to $5\%$.
One may thus wonder whether the robustness of the proposed model is due to various noise levels in the training dataset and whether the proposed model is still effective when the noise level is outside of this range.
To answer this question, we evaluate the proposed model on Levin's dataset~\cite{LevinCVPR2009} with $10\%$ Gaussian noise, a noise level that is not included in our training dataset.
The proposed model still performs comparably against the model specifically trained on images with $10\%$ Gaussian noise as shown in \cref{tab:ablation_robustness}.
In practice, this noise robustness thus allows us to avoid noise-specific training.

\begin{figure}[!t]
\scriptsize
\centering
\begin{tabular}{@{}cc@{}}
\includegraphics[width = 0.7\linewidth]{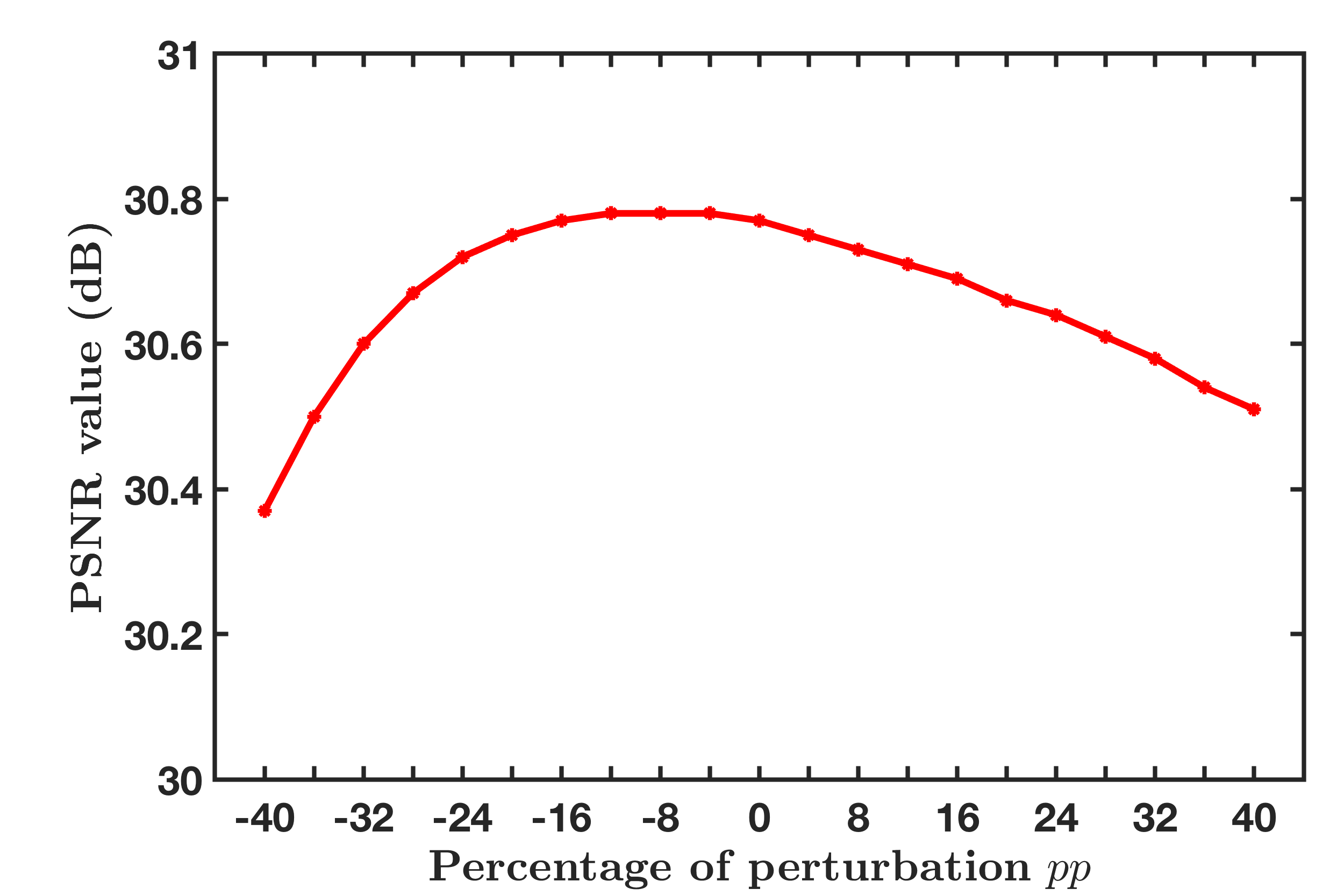}\\
\end{tabular}
\caption{Sensitivity analysis \wrt~the estimation of $\{ \frac{s_i^n}{s_i^x} \}$.
Specifically, we denote the percentage of perturbation as $pp$ and use $\{ \frac{s_i^n}{s_i^x} (1+pp) \}$ as the perturbed parameters, where $pp$ ranges from $-40\%$ to $40\%$. All PSNR values are obtained by directly applying our model trained in \cref{ssec:Results with simulated blur and Gaussian noise} with the perturbed parameters $\{ \frac{s_i^n}{s_i^x} (1+pp) \}$ (no retraining).
}
\label{fig:snr_robustness}
\end{figure}

The robustness of the proposed approach to various noise levels not only stems from the estimation of the parameters $\{ \frac{s_i^n}{s_i^x} \}$, but also from our end-to-end network.
Such benefit has also been demonstrated in high-level computer vision tasks.
Diamond \etal\ \cite{DiamondArXiv2017} propose an end-to-end architecture for joint denoising, deblurring, and classification, which makes classification more robust to realistic noise and blur.
By embedding the Wiener deconvolution into an end-to-end network, our proposed DWDN method facilitates a feature extractor for learning useful features for deconvolution with fewer artifacts.
Our architecture also benefits from feature refinement of the deconvolved features to reconstruct clearer final images.

\myparagraph{Sensitivity \wrt~the estimation of $\{ \frac{s_i^n}{s_i^x} \}$.}
As the parameters $s_i^x$ and $s_i^n$ are computed approximately but not accurately, we further carry out a sensitivity analysis \wrt~the estimation of the parameters $\{ \frac{s_i^n}{s_i^x} \}$.
We use the dataset of~\cite{LevinCVPR2009} with $5\%$ Gaussian noise and add $-40\%$ to $40\%$ perturbation to our estimated $\{ \frac{s_i^n}{s_i^x} \}$ (no retraining).
Specifically, we denote the percentage of perturbation as $pp$ and use $\{ \frac{s_i^n}{s_i^x} (1+pp) \}$ as the perturbed parameters.
\Cref{fig:snr_robustness} shows that the PSNR values differ no more than $0.06$dB within $-20\%$ to $20\%$ perturbation, suggesting the robustness of our DWDN method to the estimation of the parameters $\{ \frac{s_i^n}{s_i^x} \}$.

\begin{table}[!t]\scriptsize
\caption{Effect of training a single model for blurry images with various types of noise and outliers.
We denote the model trained on blurry images with only Gaussian noise, a mixture of blurry images degraded by either Gaussian noise or JPEG compression, and blurry images with only JPEG compression artifacts as Model$_{\text{Gaussian}}$, Model$_{\text{Mixture}}$, and Model$_{\text{JPEG compression}}$, respectively.  }
\label{tab:ablation_model_mixture}
\vspace{-1mm}
\begin{tabularx}{\linewidth}{@{}Xc@{\hspace{2.0mm}}c@{\hspace{3.0mm}}c@{\hspace{2.0mm}}c@{}}
    \toprule
~ & Noise level & Model$_{\text{Gaussian}}$ & Model$_{\text{Mixture}}$ & Model$_{\text{JPEG compression}}$ \\
    \midrule
\multirow{3}{*}{\begin{tabular}{@{}c@{}}\cite{LevinCVPR2009} with\\ Gaussian noise\end{tabular}}
&$1\%$   &36.90/0.9614  &36.83/0.9611  &--\\
&$3\%$   &32.77/0.9179  &32.76/0.9179  &--\\ %
&$5\%$   &30.77/0.8857  &30.78/0.8858  &--\\
    \midrule
\multirow{3}{*}{\begin{tabular}{@{}c@{}}\cite{MartinICCV2001} with\\$1\%$ Gaussian noise\\$\&$JPEG compression\end{tabular}}
&$50$    &--   &27.72/0.7844  &27.80/0.7877\\
&$70$    &--   &28.58/0.8167  &28.67/0.8208\\ 
&$90$    &--   &30.03/0.8607  &30.09/0.8631\\  
    \bottomrule
  \end{tabularx}
\end{table}

\begin{table}[!t]\scriptsize
\caption{Robustness to inaccurate blur kernels (PSNR in dB/SSIM). All methods are evaluated on the dataset of~\cite{LaiCVPR2016} with $1\%$ Gaussian noise, where the kernel size ranges from $51\times51$ to $101\times101$ pixels.}
\label{tab:ablation_inaccurate_kernels}
\vspace{-1mm}
\begin{tabularx}{\linewidth}{@{}Xc@{\hspace{6.0mm}}c@{\hspace{6.0mm}}cc@{}}
    \toprule
~       & ~~~~~~~~~~~~~~~~~~~~         &\multicolumn{2}{c}{Kernels for testing} \\
    \cmidrule(l){3-4}
Methods & Kernels for training           &GT  &\cite{XuECCV2010} \\
    \midrule
\cite{VasuCVPR2018}&--       &--            &23.00/0.7554\\ %
DWDN              &GT                    &27.75/0.8664  &23.74/0.7998\\
DWDN              &\cite{VasuCVPR2018}   &25.55/0.8298  &23.45/0.7834\\ %
DWDN &Mix of GT $\&$~\cite{VasuCVPR2018} &27.33/0.8620  &23.80/0.7987\\ %
    \bottomrule
  \end{tabularx}
\end{table}

\myparagraph{Single model \emph{vs.} specific model for various noise and outliers.}
In \cref{sec:experimental_results}, we have evaluated our approach on different scenarios, including blurry images with Gaussian noise, saturated pixels, or JPEG compression.
In each scenario, \eg, for blurry images with Gaussian noise, we show that a single model is able to handle various levels of Gaussian noise in each scenario, \eg, $1\%$--$5\%$.
In this section, we additionally evaluate the effect of training a single model for blurry images with \emph{varying} types of noise and outliers. 
To do this, we train a model using the same clear training images as in \cref{ssec:Results with simulated blur and Gaussian noise}, where each blurry image is degraded by either Gaussian noise of $1\%$--$5\%$ or JPEG compression with the quality parameter of \num{50}--\num{90} (\emph{Model$_{\text{Mixture}}$} for short).
\Cref{tab:ablation_model_mixture} shows that using a single model trained on mixed blurry images with Gaussian noise or JPEG compression artifacts generates similar results compared to using the one trained on images with a specific type of noise or outliers (\emph{Model$_{\text{Gaussian}}$}, \ie, DWDN in \cref{tab:levin_dataset}, and \emph{Model$_{\text{JPEG compression}}$}, \ie, DWDN in \cref{tab:bsds_dataset_jpeg} for short).
This robustness makes it possible to learn a single model that can handle various types of noise and outliers.

\myparagraph{Robustness to inaccurate kernels.}
Blur kernels estimated by blind image deblurring methods usually contain substantial errors.
Thus, it is necessary to evaluate the robustness of our proposed non-blind method to inaccurate kernels.
We use the same method as described in~\cref{ssec:Results with simulated blur and Gaussian noise} to generate the training data.
To train our model for this task, we respectively use ground-truth (GT) kernels, synthetic noisy kernels by~\cite{VasuCVPR2018}, and a mix of GT and synthetic noisy kernels.
We then compare with the method of~\cite{VasuCVPR2018}, which focuses on handling inaccurate kernels and use the same test dataset~\cite{LaiCVPR2016} as~\cite{VasuCVPR2018} for evaluation.
The dataset~\cite{LaiCVPR2016} contains $1\%$ Gaussian noise and the kernel size ranges from $51\times51$ to $101\times101$ pixels.
\Cref{tab:ablation_inaccurate_kernels} shows that the proposed approach is robust to inaccurate kernels, even when trained with GT kernels, improving the results over \cite{VasuCVPR2018} by a large margin of $0.7$dB.

Compared to competing approaches, our method performs robustly to noise and inaccurate kernels for two reasons.
First, benefitting from the proposed feature-based Wiener deconvolution, our method can adaptively estimate the noise-related parameters $s_i^x$ and $s_i^n$ as stated in~\cref{ssec:feature_deconvolution}.
Second, our end-to-end network facilitates the feature extractor for learning useful features for deconvolution with fewer artifacts, as well as benefits from the feature refinement module that uses redundant deconvolved features to reconstruct clearer images.
The robustness is also evaluated on real-world images as shown  in~\cref{fig:motivation1,fig:visual_results_real1}.
More comparisons are included in the supplemental material.

\begin{figure}[t]
\scriptsize\sffamily
\centering
\begin{tabular}{@{\hspace{-1mm}}c@{\hspace{1mm}}c@{\hspace{1mm}}c@{\hspace{1mm}}c@{}}
\includegraphics[width = 0.33\linewidth]{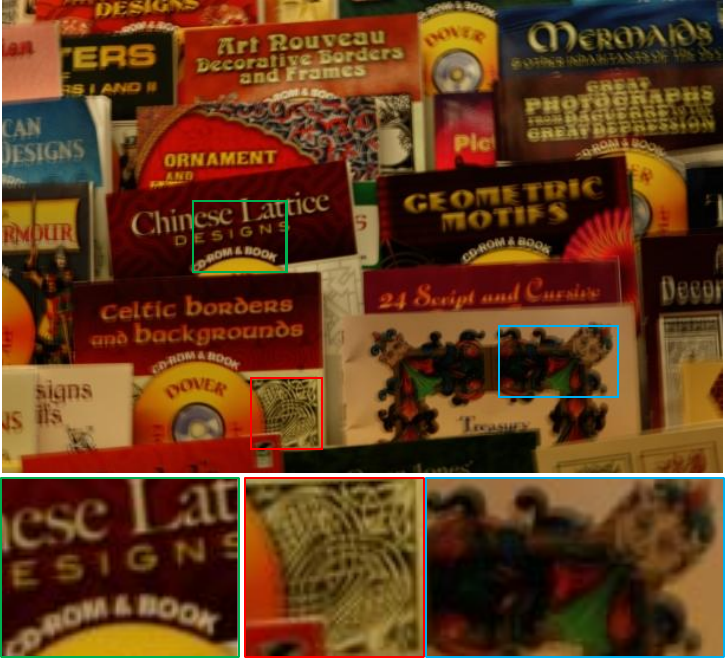}&
\includegraphics[width = 0.33\linewidth]{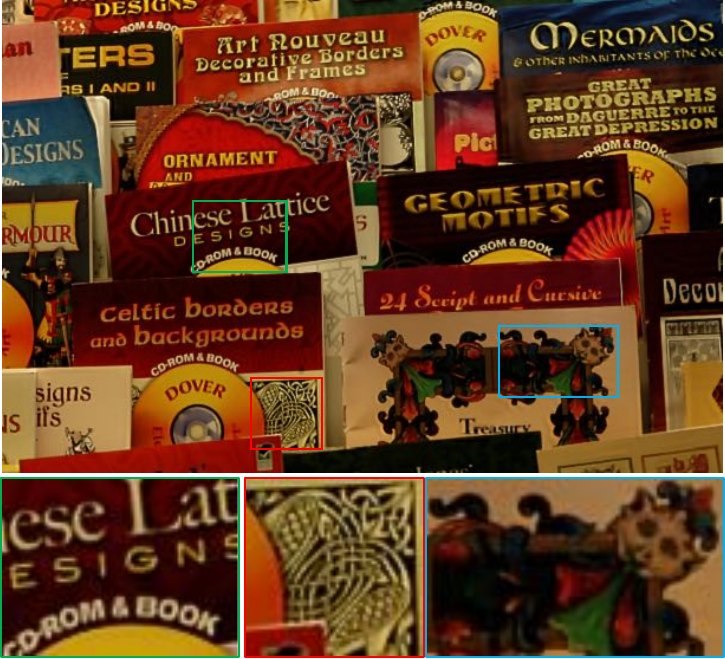}&
\includegraphics[width = 0.33\linewidth]{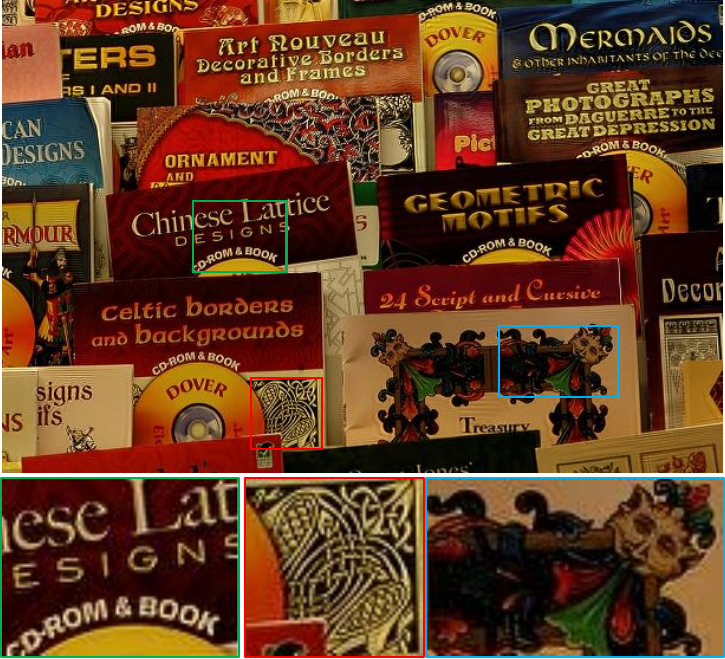} \\
\includegraphics[width = 0.33\linewidth]{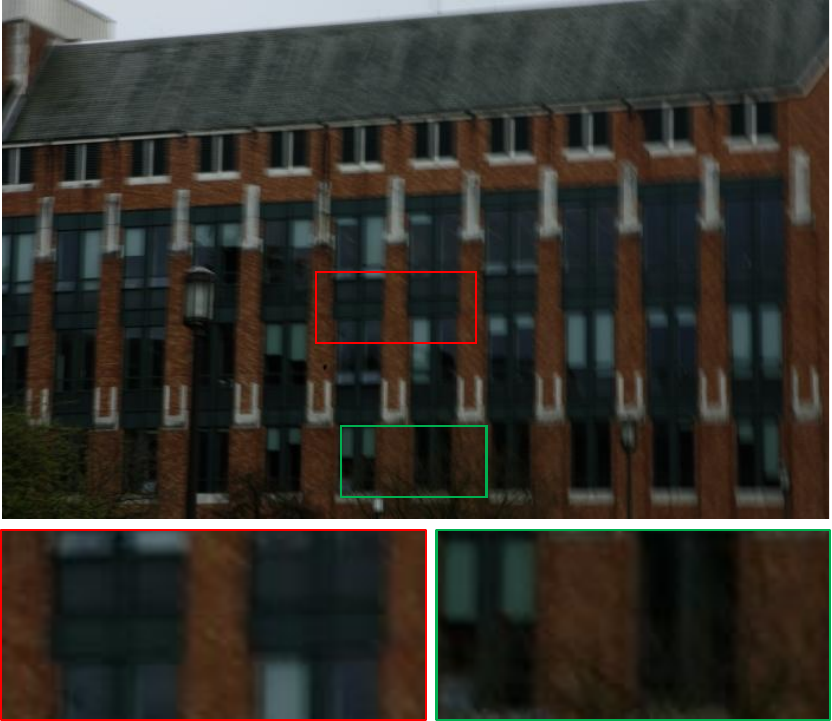}&
\includegraphics[width = 0.33\linewidth]{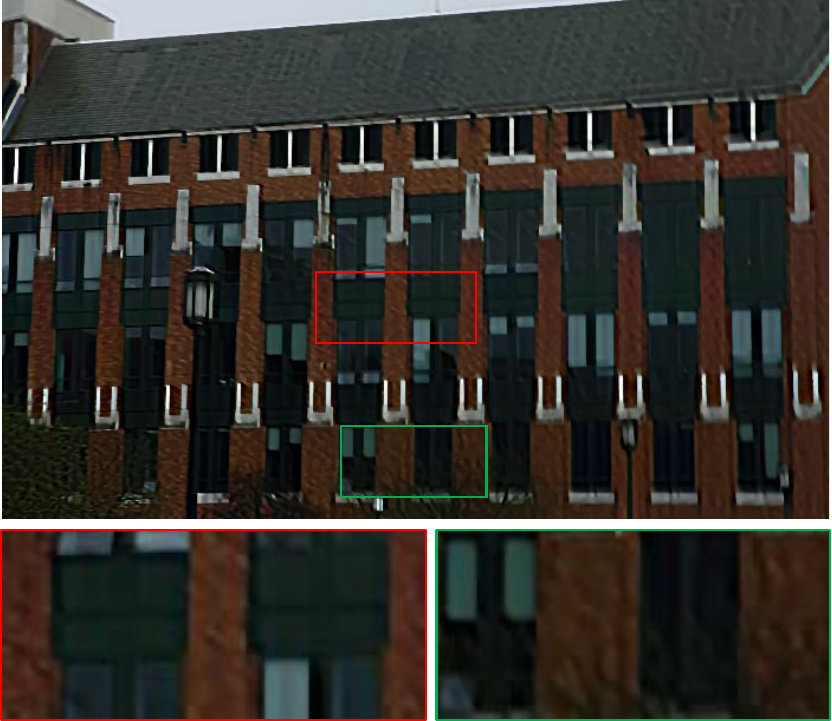}&
\includegraphics[width = 0.33\linewidth]{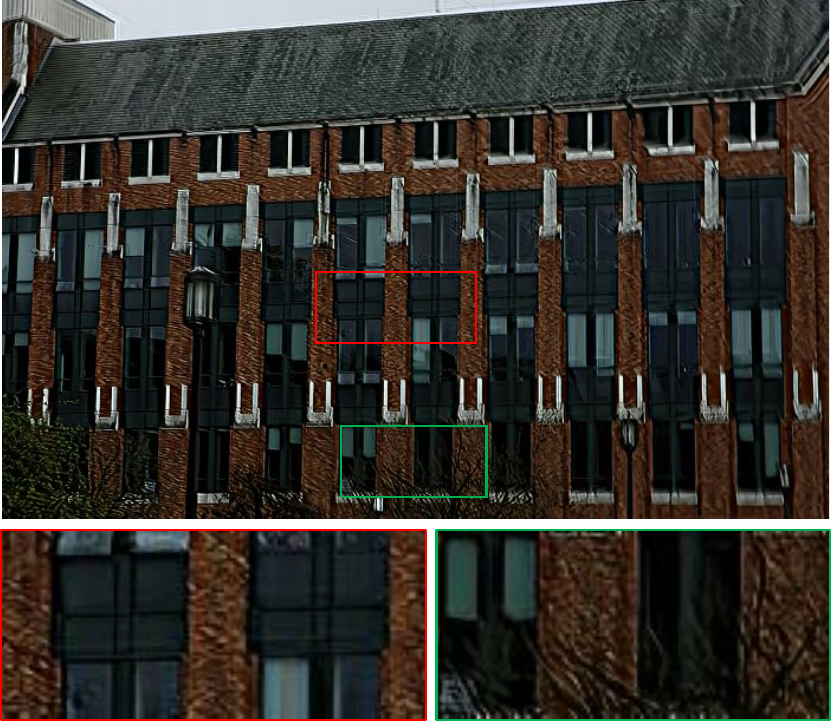} \\
(a) Blurry input & (b) \cite{PanPAMI2017} & (c) DWDN+ \\
\end{tabular}
\caption{Examples with non-uniform image deblurring. The images shown in \emph{(b)} are obtained from the reported results of~\cite{PanPAMI2017}. Compared to the results in \emph{(b)}, the images recovered by our method in \emph{(c)} are notably clearer with finer detail.
}
\label{fig:non-uniform}
\end{figure}

\subsection{Extension to non-uniform image deblurring}
\label{ssec:Extension to non-uniform image deblurring}
Our method is derived based on a uniform blur model.
However, our method can be extended to handle non-uniform blur by applying the local uniform approximation method of~\cite{WhyteIJCV2014}.
We evaluate the effectiveness of our method on non-uniform image deblurring in~\cref{fig:non-uniform}, where the blurry examples are from~\cite{GuptaECCV2010} and the blur kernels are estimated by the method of~\cite{PanPAMI2017}.
The result by~\cite{PanPAMI2017} over-smoothes the detail in the restored images, as shown in~\cref{fig:non-uniform}(b).
In contrast, our method generates visibly clearer images in~\cref{fig:non-uniform}(c), where the fine-scale structures of the books and branches are recovered much better.

\subsection{Run-time}
\label{ssec:Run-time}
We benchmark the run-time of a selection of evaluated methods on a machine with an Intel Xeon E5-2650 v4 CPU and an NVIDIA TITAN Xp GPU.
\Cref{tab:running_time} summarizes the average run-time of representative methods.
Our DWDN approach requires about $0.05$ seconds on the images (with $255\times255\times3$ pixels) from Levin \etal~\cite{LevinCVPR2009} and roughly $0.40$ seconds on images (with $800\times1024\times3$ pixels) from Sun \etal~\cite{SunICCP2013}.
The methods~\cite{ZhangCVPR2017,KruseICCV2017,GongTNNLS2020} are iterative and thus take more time.
The approach of~\cite{SonICCP2017} contains a postprocessing step of solving an optimization problem, which also takes a certain amount of time.
In contrast, the proposed deep Wiener deconvolution network is based on an end-to-end architecture, which does not require iterative solutions or postprocessing steps.
Thus, the proposed DWDN approach runs faster with high image quality.
Our improved DWDN+ method employs a cascade architecture, which takes more time than DWDN.
However, the run-time of DWDN+ is still comparable or even less than competing methods, highlighting the efficiency of our proposed approach.

\begin{table}[!t]
\scriptsize
  \caption{Run-time performance (seconds). All methods are evaluated on the same machine with the same settings.}
  \vspace{-1mm}
  \label{tab:running_time}
  \centering
  \begin{tabularx}{\linewidth}{@{}Xc@{\hspace{0.5mm}}c@{\hspace{1mm}}c@{\hspace{1mm}}c@{\hspace{0.5mm}}c@{\hspace{0.5mm}}c@{}}
    \toprule
Image size & IRCNN~\cite{ZhangCVPR2017} & FDN~\cite{KruseICCV2017} & FNBD~\cite{SonICCP2017} & RGDN~\cite{GongTNNLS2020} & DWDN &DWDN+\\
    \midrule
$255\times255\times3$   &0.32 &0.60 &0.07 &1.44  &0.05 & 0.13\\
$800\times1024\times3$  &4.59 &1.85 &0.67 &11.27 &0.40 & 0.97\\
    \bottomrule
  \end{tabularx}
\end{table}

\begin{figure}[t]
\scriptsize\sffamily
\centering
\begin{tabular}{@{\hspace{-1mm}}c@{\hspace{1mm}}c@{\hspace{1mm}}c@{\hspace{1mm}}c@{}}
\includegraphics[height = 2.2cm]{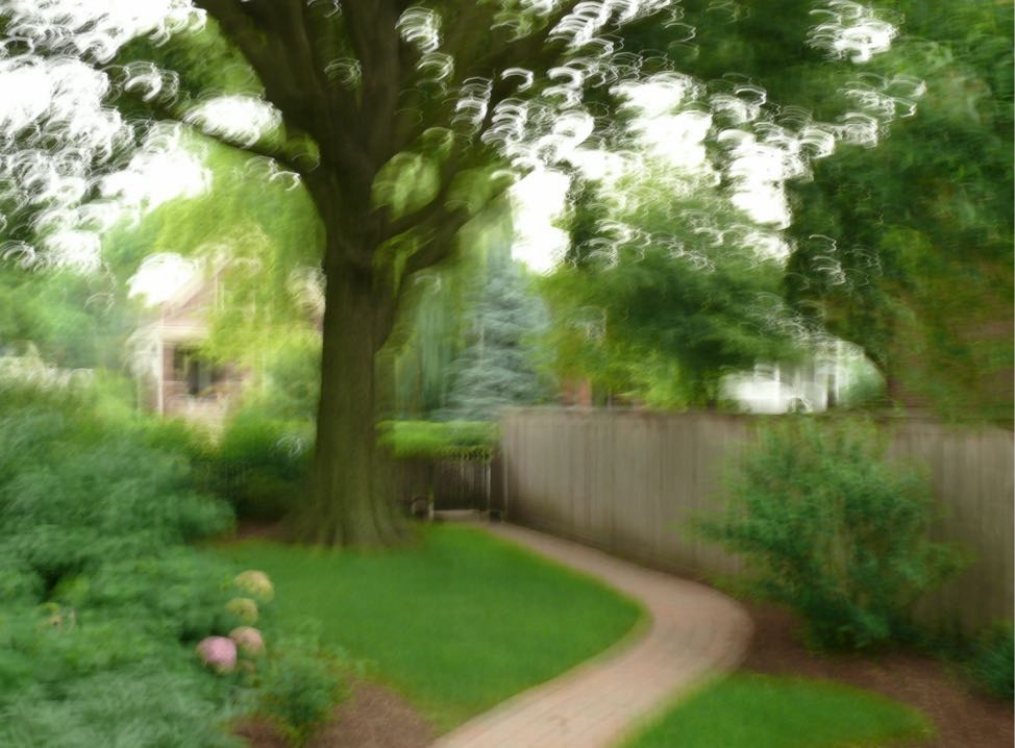}&
\includegraphics[height = 2.2cm]{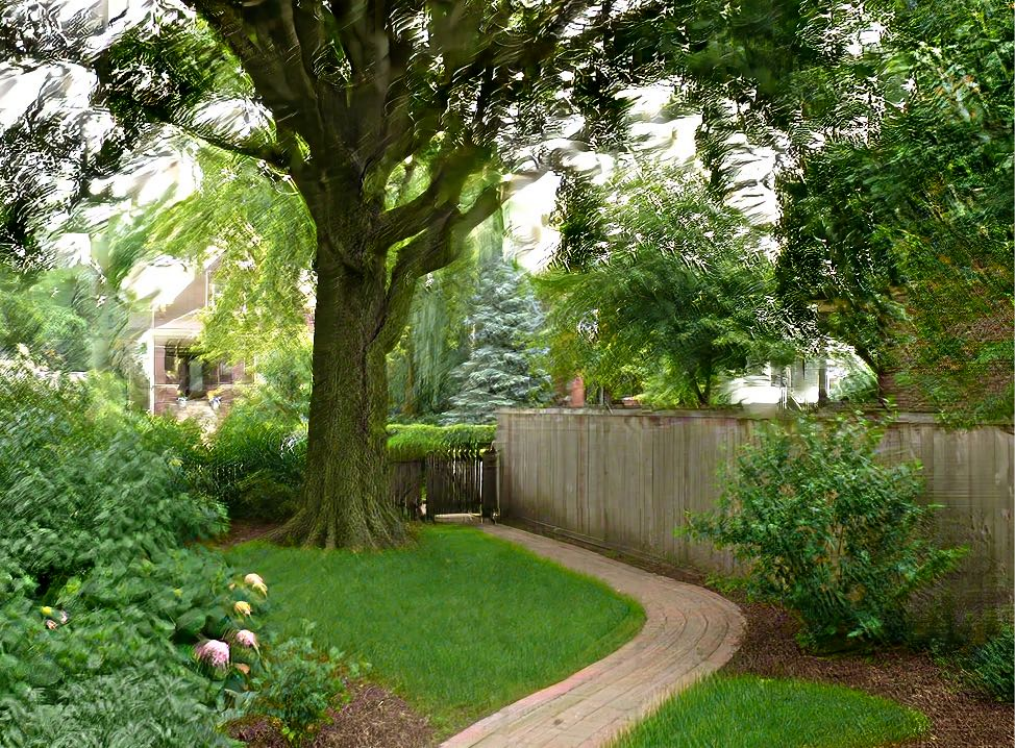}&
\includegraphics[height = 2.2cm]{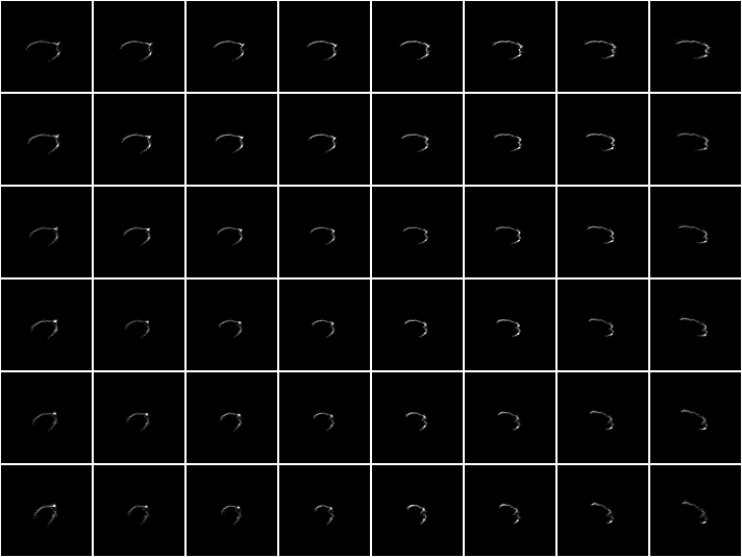}\\
(a) Blurry input & (b) DWDN+ & (c) Estimated kernel \\
\end{tabular}
\caption{A failure case on a challenging image from \cite{WhyteIJCV2014}. The spatially-variant blur kernels in the blurry image \emph{(a)} are estimated by the method of Whyte \etal~\cite{WhyteIJCV2012,WhyteIJCV2014}, as shown in \emph{(c)}. Our approach does not effectively recover a clear image in \emph{(b)}, especially in the top half of the image, which corresponds to the region that contains large blur and a great number of saturated areas in the blurry input \emph{(a)}.
}
\label{fig:limitation}
\end{figure}

\subsection{Limitations}
\label{ssec:Limitations}
Our approach focuses on handling the non-blind deconvolution process, not including the kernel estimation process. 
Thus, the performance of the proposed method on real images is affected by the accuracy of the blur kernel estimation.
Although our approach is empirically more robust to inaccurate kernels than e.g.~\cite{VasuCVPR2018} in \cref{tab:ablation_inaccurate_kernels}, our results with estimated kernels are clearly worse than those with GT kernels.
This demonstrates that there is room for improvement in non-blind image deblurring with inaccurate kernels.

Furthermore, the proposed method may fail when the blurry image contains a large number of saturated areas or severe blur. 
As we do not have a specific design for handling large saturated regions and the blur kernel estimated in these cases can be faulty, our approach may not be able to effectively recover a clear image (\cref{fig:limitation}(b)).
In future work, we aim to incorporate the error in kernel estimation and outlier modeling for non-blind deconvolution to improve the robustness of non-blind image deblurring.

\section{Conclusion}
In this paper, we propose a feature-based Wiener deconvolution module, in which we explore useful learned features from deep neural networks to yield an explicit Wiener deconvolution step in the extracted deep feature space.
We show that compared with the commonly used deconvolution conducted in the image space, it is much more effective to deconvolve the blurry image in a (deep) feature space. 
We further develop a multi-scale cascaded feature refinement module to progressively restore fine-scale detail from the deconvolved features.
Extensive evaluations and comparisons with state-of-the-art methods demonstrate that our approach extends the state of the art in non-blind deblurring by a wide margin while being robust to different noise levels and inaccurate blur kernels.

\appendices



\ifCLASSOPTIONcompsoc
  \section*{Acknowledgments}
\else
  \section*{Acknowledgment}
\fi

This project has received funding from the European Research Council (ERC) under the European Union's Horizon 2020 research and innovation programme (grant agreement No.~866008).
We thank Uwe Schmidt for helpful feedback on the manuscript.

\ifCLASSOPTIONcaptionsoff
  \newpage
\fi



\bibliographystyle{IEEEtran}
\bibliography{egbib}
%



%

\begin{IEEEbiography}[{\includegraphics[width=1in,clip]{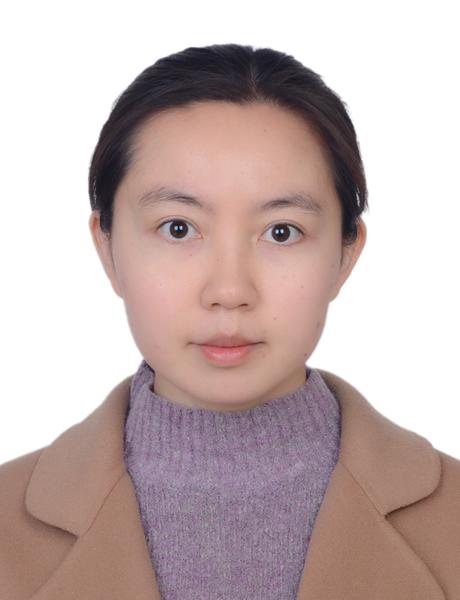}}]{Jiangxin Dong}
received the Diploma degree in Information and Computational Science from Dalian University of Technology in 2014 and the PhD degree in Computational Mathematics from Dalian University of Technology in 2019. She was a postdoctoral researcher in the Department of Computer Vision and Machine Learning, Max Planck Institute for Informatics, Germany. She is currently a professor with Nanjing University of Science and Technology, China. Her research interests include image restoration, especially image deblurring.
\end{IEEEbiography}

\begin{IEEEbiography}[{\includegraphics[width=1in,clip]{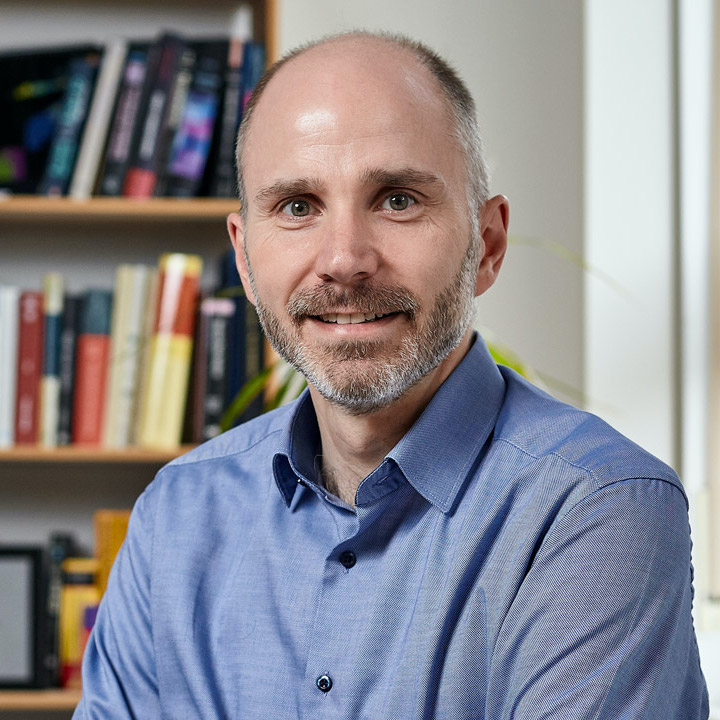}}]{Stefan Roth}
received the Diploma degree in Computer Science and Engineering from the University of Mannheim, Germany in 2001. In 2003 he received the ScM degree in Computer Science from Brown University, and in 2007 the PhD degree in Computer Science from the same institution. Since 2007, he is on the faculty of Computer Science at Technische Universität Darmstadt, Germany, where is currently full professor (W3). His research interests include learning approaches to image modeling, motion estimation and tracking, as well as scene understanding. He received several awards, including the Olympus-Prize 2010 of the German Association for Pattern Recognition (DAGM), the Heinz Maier-Leibnitz Prize 2012 of the German Research Foundation (DFG), the Longuet-Higgins Prize 2020, as well as various paper awards. He is the recipient of two grants from the European Research Council (ERC), a Starting Grant in 2013 and a Consolidator Grant in 2020. He is a Fellow of the European Laboratory for Learning and Intelligent Systems (ELLIS) and directs the ELLIS Unit at TU Darmstadt. He regularly serves as an area chair for CVPR, ICCV, and ECCV, and will be program co-chair of CVPR 2022 and ECCV 2024. 
\end{IEEEbiography}

\begin{IEEEbiography}[{\includegraphics[width=1in,height=1.25in,clip,keepaspectratio]{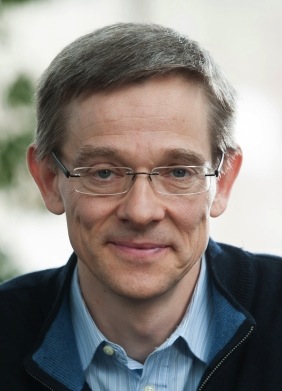}}]{Bernt Schiele} received masters degree in computer science from the University of Karlsruhe and INP Grenoble in 1994 and the PhD degree from INP Grenoble in computer vision in 1997. He was a postdoctoral associate and visiting assistant professor with MIT between 1997 and 2000. From 1999 until 2004, he was an assistant professor with ETH Zurich and, from 2004 to 2010, he was a full professor of computer science with TU Darmstadt. In 2010, he was appointed a scientific member of the Max Planck Society and director at the Max Planck Institute for Informatics. Since 2010, he has also been a professor at Saarland University. He is particularly interested in developing methods which work under real world conditions.
\end{IEEEbiography}  




\end{document}